\documentclass[10pt,twocolumn,letterpaper]{article}

\usepackage[pagenumbers]{iccv} 

\usepackage{times}
\usepackage{xspace}
\usepackage{epsfig}
\usepackage{graphicx}
\usepackage{xcolor}
\usepackage{lipsum}
\usepackage[export]{adjustbox}
\usepackage{pifont}
\usepackage{amsmath}
\usepackage{makecell}
\usepackage{multirow}
\usepackage{colortbl}

\newcommand{\ChartCap}{\textsc{ChartCap}\xspace}
\newcommand{\cmark}{\ding{51}}
\newcommand{\xmark}{\ding{55}\xspace}
\definecolor{darkgreen}{rgb}{0.0, 0.55, 0.0}
\definecolor{lightgray}{gray}{0.85}

\definecolor{iccvblue}{rgb}{0.21,0.49,0.74}
\usepackage[pagebackref,breaklinks,colorlinks,allcolors=iccvblue]{hyperref}

\def\paperID{15519} 
\def\confName{ICCV}
\def\confYear{2025}

\title{\ChartCap: Mitigating Hallucination of Dense Chart Captioning}

\author{Junyoung Lim \quad Jaewoo Ahn \quad Gunhee Kim\\
Seoul National University\\
{\tt\small \{junyoung.lim, jaewoo.ahn\}@vision.snu.ac.kr, gunhee@snu.ac.kr}\\
{\tt\small \href{https://junyoung-00.github.io/ChartCap/}{https://junyoung-00.github.io/ChartCap/}}\
}

\begin{document}

\maketitle

\begin{abstract}
Generating accurate, informative, and hallucination-free captions for charts remains challenging for vision language models, primarily due to the lack of large-scale, high-quality datasets of real-world charts. However, existing real-world chart datasets suffer from the inclusion of extraneous information that cannot be inferred from the chart and failure to sufficiently capture structural elements and key insights. Therefore, we introduce ChartCap, a large-scale dataset of 565K real-world chart images paired with type-specific, dense captions that exclude extraneous information and highlight both structural elements and key insights in detail. To build ChartCap, we design a four-stage pipeline that generates captions using only the discernible data from the chart and employ a cycle consistency-based human verification, which accelerates quality control without sacrificing accuracy. Additionally, we propose a novel metric, the Visual Consistency Score, which evaluates caption quality by measuring the similarity between the chart regenerated from a caption and the original chart, independent of reference captions. Extensive experiments confirms that models fine-tuned on ChartCap consistently generate more accurate and informative captions with reduced hallucinations, surpassing both open-source and proprietary models and even human-annotated captions.
\end{abstract}    

\section{Introduction}
\label{sec:intro}
Charts are powerful tools for visualizing data distributions, trends, and patterns across various domains such as science, economics, and sociology. By presenting complex information in a concise and intuitive manner~\citep{larkin1987diagram, pandey2014persuasive}, charts help readers gain meaningful insights for decision-making process. However, charts involve complex spatial relationships among various elements such as axes, labels, and legends, which can be interpreted differently depending on the chart type. Consequently, it is challenging not only for humans but also for  vision language models (VLMs) to interpret complex charts~\citep{carberry2006information, kim2020answering, stokes2022striking}.

Chart captioning is one of the core tasks to assess VLMs’ ability to understand charts. Its goal is to generate natural descriptions of a chart image~\citep{kantharaj2022chart}. An ideal caption should (1) avoid inaccuracies about the chart~\citep{lo2022misinformed, huang2023lvlms, akhtar2024chartcheck}, and (2) include a structural description of the chart components (e.g., title or legends) as well as key insights such as major statistics (e.g., maximum or minimum values) and perceptual patterns (e.g., data trends)~\citep{lundgard2021accessible, tang2023vistext}. However, existing real-world chart datasets~\citep{hsu2021scicap, kantharaj2022chart, rahman2023chartsumm, liu2023mmc, li2024multimodal}  suffer from two major issues: (1) they contain extraneous information in captions that cannot be inferred from the image, and (2) they fail to sufficiently capture the essential information specific to each chart type.

\begin{figure*}[t]
\centering
\includegraphics[width=0.90\textwidth]{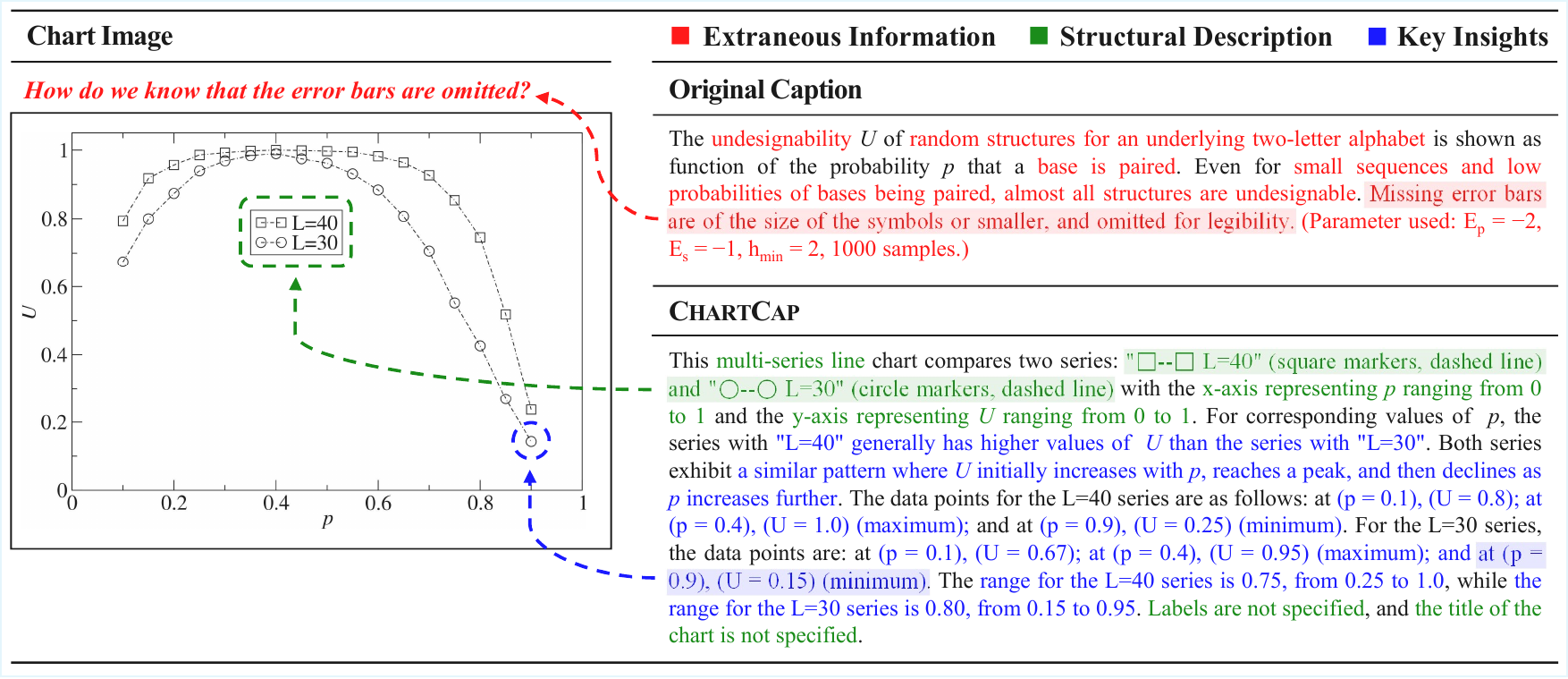}  
\caption{Comparison of the original caption and our \ChartCap caption. The original caption includes extraneous information (in \textcolor{red}{red}), such as additional contextual details (e.g., missing error bars) and references to parameters (\(E_p\), \(E_s\), \(h_{\text{min}}\)), which cannot be inferred from the chart image. In contrast, \ChartCap caption follows the line chart schema, relying on the information visible in the image. It includes a structural description (in \textcolor{darkgreen}{green}) and key insights (in \textcolor{blue}{blue}). The chart  is sourced from~\citep{Burghardt_2007}, collected by~\citep{li2024multimodal}, and included in \ChartCap.}
\label{fig:chartcap_example}
\end{figure*}

First, the datasets contain \textit{extraneous information} within their captions, 
mainly because charts are usually embedded in source documents and their original captions are simply paired with chart images without verification.
Captions are often written based not only on the chart itself but also on the surrounding context. As a result, these captions include the information that cannot be inferred from the chart image alone (but from text in the document together), as shown in Figure~\ref{fig:chartcap_example}. This poses an ill-posed problem for expecting the model to predict information absent from the chart, ultimately leading to hallucination. 

Second, real-world chart datasets lack sufficient structural description and key insights in text; they often omit critical information that the chart image conveys (see Figure~\ref{fig:chartcap_example}). It is partly because authors do not specify some details in the captions, assuming that human readers can easily infer them from the figure~\citep{carberry2006information, masry2023unichart}.
Such information varies depending on the chart type; for example, scatter plots highlight clusters and distributions, while line charts emphasize temporal trends and changes~\citep{cleveland1985elements}. Hence, a \textit{type-specific caption schema}, which specifies how to interpret critical information for each chart type, is required to enable models to generate informative captions.

To address these issues, we propose \ChartCap to improve VLMs’ captioning performance while mitigating hallucinations. \ChartCap is a large-scale dataset of real-world chart images, containing 565K chart-caption pairs that (1) exclude extraneous information not verifiable from the chart image and (2) provide structural description and key insights in a dense manner by following a type-specific caption schema. Drawing on research in the data visualization domain~\citep{munzner2014visualization, lee2016vlat}, we define a caption schema that structures the core information to be included for each chart type. We then devise an automatic pipeline that generates captions using only the data inherent in each chart image, thereby minimizing the inclusion of extraneous information. Finally, we employ a cycle consistency-based~\citep{zhu2017unpaired, pesaran-zadeh-etal-2024-text2chart31} human verification to ensure high-quality data pairs. Figure~\ref{fig:chartcap_example} illustrates a comparative example between \ChartCap and the original captions. 

Moreover, we propose a reference-free metric, the \textit{Visual Consistency Score} (VCS), for evaluating chart captions. VCS exploits a recently powerful large language model (LLM) that translates a caption into Python code to generate a chart. Then, it compares the reconstructed chart to the ground-truth chart, overcoming the limitations of existing automated metrics, which struggle to capture the deep semantic quality of captions and are highly dependent on the quality of reference captions. In a head-to-head study, VCS demonstrated high agreement rate with human judgments, outperforming existing automatic metrics such as BERTScore~\citep{zhang2019bertscore}.

Extensive experiments show that VLMs fine-tuned on \ChartCap consistently generate more informative captions with fewer hallucinations, in terms of reference-based metrics, human evaluation, and the Visual Consistency Score, surpassing both open-source and proprietary models, including InternVL2.5~\citep{chen2024expanding},  Phi3.5-vision~\citep{abdin2024phi}, ChartGemma~\citep{masry2024chartgemma}, ChartInstruct-Llama2~\citep{masry2024chartinstruct} and Claude 3.5 Sonnet~\citep{claude3.5}. Moreover, the captions generated by \ChartCap fine-tuned VLMs are more preferred to human-annotated captions from VisText~\citep{tang2023vistext} and Chart-to-Text~\citep{kantharaj2022chart} by human evaluators. 

In summary, our main contributions are as follows:
\begin{enumerate}
    \item We propose \ChartCap, a large-scale  565K real-world chart caption dataset that is free from extraneous information and correctly conveys structural description and key insights via type-specific caption schema.
    \item We propose the Visual Consistency Score (VCS), which evaluates the quality of chart captions by assessing deep semantic meaning without relying on reference captions.
    \item Through extensive experiments, we show that VLMs trained on \ChartCap generate high-quality, informative captions with fewer hallucinations.
\end{enumerate}

{\renewcommand{\arraystretch}{1.0}
\begin{table}[t!] 
\centering
\begin{adjustbox}{width=\columnwidth}
\begin{tabular}{lcccccccc}
    \toprule
    \makecell[l]{\textbf{Dataset}} & \makecell{\textbf{Real-world}\\\textbf{charts}} & \makecell{\textbf{Free from }\\\textbf{extraneous info}} & \makecell{\textbf{Type-specific}\\\textbf{schema}} & \makecell{\textbf{Human}\\\textbf{annotation}} & \makecell{\textbf{Data}\\\textbf{pairs}} \\
    \midrule
    \makecell[l]{ChartLlama~\cite{han2023chartllama}} & \xmark & \cmark & \xmark & \xmark & 11K\\
    \makecell[l]{VisText~\cite{tang2023vistext}} & \xmark & \cmark & \cmark & 12K & 12K \\
    \makecell[l]{AutoChart~\cite{zhu2021autochart}} & \xmark & \cmark & \cmark & \xmark & 24K\\
    \midrule
    \makecell[l]{ChartGemma~\cite{masry2024chartgemma}} & $\Delta$ & \cmark & \xmark & \xmark & 62K \\
    \makecell[l]{ChartSFT~\cite{meng2024chartassisstant}} & $\Delta$ & \xmark & \xmark & \xmark & 1.0M \\
    \makecell[l]{MMC~\cite{liu2023mmc}} & $\Delta$ & \xmark & \xmark & \xmark & 400K \\
    \midrule
    \makecell[l]{ArxivCap~\cite{li2024multimodal}} & \cmark & \xmark & \xmark & \xmark & 3.9M* \\
    \makecell[l]{ChartSumm~\cite{rahman2023chartsumm}} & \cmark & \xmark & \xmark & \xmark & 84K \\
    \makecell[l]{Chart-to-Text~\cite{kantharaj2022chart}} & \cmark & \xmark & \xmark & 8K & 44K \\
    \makecell[l]{SciCap~\cite{hsu2021scicap}} & \cmark & \xmark & \xmark & \xmark & 134K \\
    \midrule
    \makecell[l]{\ChartCap} & \cmark & \cmark & \cmark & \textbf{56K} & \textbf{565K} \\
    \toprule
\end{tabular}
\end{adjustbox}
\caption{Comparison of \ChartCap with public chart captioning datasets. Datasets marked with $\Delta$ include both real-world and synthetic charts. The asterisk (*) for ArxivCap indicates that it comprises both data-driven charts and non-data-driven ones such as conceptual diagrams or scientific illustrations. The Human annotation column means the number of chart-caption pairs annotated or verified by human. \ChartCap encompasses 565K real-world chart-captions with human verification applied on the test set.}
\label{tab:chart_comparison}
\end{table}}

\section{Related Work}
\subsection{Datasets}
For VLMs to generate accurate and informative chart captions, the training data should consist of real chart images, contain correct information, and capture the key insights conveyed by the chart.
While synthetic datasets are scalable, models trained on synthetic data tend to exhibit limited robustness when applied to real-world charts \cite{xu2023chartbench, zeng2024advancing}. AutoChart~\cite{zhu2021autochart}, VisText~\cite{tang2023vistext}, ChartLlama~\cite{han2023chartllama}, and ChartSFT~\cite{meng2024chartassisstant} are generated programmatically from raw data using visualization tools. However, ChartBench~\cite{xu2023chartbench} shows that LLaVA~\cite{liu2024improved} trained on synthetic ChartLlama~\cite{han2023chartllama} underperforms its pre-training baseline. 

In contrast, ChartInstruct~\cite{masry2024chartinstruct} collects real charts from 157  websites. However, it remains inaccessible to the public due to legal constraints. ChartSumm~\cite{rahman2023chartsumm} and Chart-to-Text~\cite{kantharaj2022chart} collect chart-caption pairs from Statista, Pew, and Knoema, but are relatively small and lack informativeness~\cite{masry2023unichart}. ChartGemma~\cite{masry2024chartgemma} leverages Gemini 1.5 Flash to regenerate captions from chart images via zero-shot prompting. MMC-Instruct~\cite{liu2023mmc}, SciCap~\cite{hsu2021scicap}, and ArxivCap~\cite{li2024multimodal} use scientific papers on arXiv, resulting in larger datasets, but model-generated captions are reported to be highly hallucinated~\cite{li2024multimodal}. 

On the other hand, our \ChartCap leverages the visual diversity of real-world charts while excluding extraneous information and utilizing caption schema, thereby enabling VLMs to acquire more robust chart comprehension skills.
More systematic comparison is presented in Table \ref{tab:chart_comparison}.

\subsection{Automatic Evaluation Metrics}
Various automatic evaluation approaches have been popularly used to measure the quality of generated captions, including BLEU~\citep{papineni2002bleu}, ROUGE~\citep{lin2004rouge}, METEOR~\citep{banerjee2005meteor}, and  BERTScore~\citep{zhang2019bertscore}, to name a few.

Despite their widespread use, these automatic evaluation metrics share common limitations. First, they fail to capture deeper linguistic or semantic nuances, captions are measured by aligning words or short phrases — even if the generated text contains factual errors or incoherent logic. Second, they are highly dependent on the quality of reference captions. Even if a generated caption accurately describes an image, it may be unfairly penalized if the reference caption is inaccurate or overly concise. Fundamentally, the true \textit{ground-truth} (GT) in image captioning is the image itself, yet existing automatic metrics do not directly compare captions to the visual content of the image.

CLIPScore~\citep{hessel2021clipscore} utilizes CLIP~\citep{radford2021clip} to directly compute the semantic similarity between an image and its caption. However, it primarily measures high-level semantic alignment~\citep{li2023evaluating} and cannot handle long captions, limiting its reliability as a comprehensive evaluation metric for tasks that require precise and detailed descriptions.

To address these challenges, we introduce a metric, \textbf{Visual Consistency Score}, which evaluates a generated caption by reconstructing the chart and computing the similarity between the reconstructed chart and the GT chart.

\subsection{Hallucinations in VLMs}
Hallucination in VLMs refers to the instances where the model generates text that does not align with the visual content~\cite{rohrbach2018object}.
One of the primary causes of hallucination is the misalignment between vision and language modalities~\cite{wang2024mitigating}. To address this, Ciem~\citep{hu2023ciem} and ~\citet{jiang2024hallucination} employ contrastive learning with carefully crafted question-answer pairs that push misaligned representations away from correct ones. \citet{liu2023mitigating} propose containing both positive and negative instructions to strengthen model robustness. \citet{sun2023aligning} and RLHF-V~\citep{yu2024rlhf} refine the training process using human feedback to reward factual outputs, while HA-DPO~\citep{zhao2023beyond}, FDPO~\citep{gunjal2024detecting}, and CLIP-DPO~\citep{ouali2025clip} leverage preference optimization by ranking and filtering generated responses.

However, they address object-centric tasks, leaving chart domain relatively unexplored. \ChartCap tightly couples textual and visual cues in chart interpretation, enabling models trained on it to exhibit fewer hallucinations in more data-driven, abstract scenarios.

\begin{figure*}[t]
\begin{center}
\vspace{-12pt}
\includegraphics[width=0.90\textwidth]{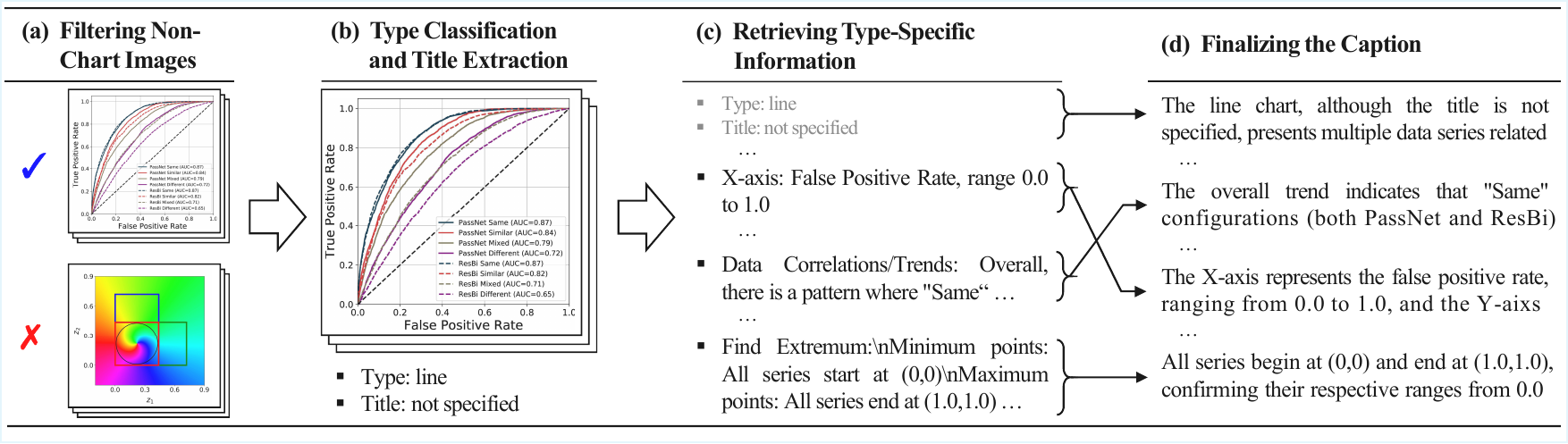}  
\end{center}
\vspace{-12pt}
\caption{An example of the four-stage pipeline for our \ChartCap: (a) filtering non-chart images, (b) classifying the chart type and extracting titles, (c) retrieving structural components and key insights, and (d) transforming the accumulated information into a coherent, sentence-level caption.}
\label{fig:pipeline_figure}
\end{figure*}

\section{The \ChartCap Dataset}
Building a large-scale chart dataset with informative captions presents several challenges. First, it requires a clear definition of what information should be included in each caption. Second, an automated procedure with an appropriate schema is needed to generate high-quality captions at minimum cost. Therefore, we define a type-specific caption schema and a caption generation pipeline with four phases. Additionally, we facilitate efficient human verification using cycle consistency \citep{zhu2017unpaired, pesaran-zadeh-etal-2024-text2chart31}, comparing the original chart image against a reconstructed image, enabling effective quality control of the test set.

\textbf{The Chart Corpora}. To assemble real-world charts, we collect 3.1 million chart images from ArxivCap~\citep{li2024multimodal}, ChartSumm-Knoema~\citep{rahman2023chartsumm}, ChartCheck~\citep{akhtar2024chartcheck}, and ChartQA-train~\citep{masry2022chartqa} as a pool of data.

\subsection{Defining the Caption Schema}
We define a \textbf{type-specific caption schema} that outlines the structural description and key insights for each of nine chart types, including \textit{line, bar, pie, histogram, scatter, area, bubble, choropleth map, and treemap}, guided by the prior work in the field of data visualization.
As a reference, \textit{Visualization Analysis and Design}~\citep{munzner2014visualization} provides a rigorous framework for designing visual representations.

To define the key insights for each chart type, we leverage the test blueprint from the \textit{Visualization Literacy Assessment Test} (VLAT)~\citep{lee2016vlat}, which identifies  cognitive tasks for non-expert readers. Based on this framework, we minimize the ambiguity inherent in the criteria for crafting informative, high-quality captions~\citep{piantadosi2012communicative, lundgard2021accessible}. The complete schema is detailed in Appendix \ref{appendix:type-specific_caption_schema}.

\subsection{Automated Dataset Generation Pipeline}
We develop a four-stage pipeline, as depicted in Figure \ref{fig:pipeline_figure}, to automate caption generation while balancing accuracy and computational cost via a combination of open-source and proprietary models. We report the accuracy of each stage by manual inspection on 100 randomly sampled instances.

\textbf{Filtering Non-Chart Images.}
We first employ InternVL2.5-8B ~\citep{chen2024expanding} to filter out non data-driven chart images (e.g., diagrams, schematics, illustrations). During this phase, multi-chart images are also removed, leaving us with 1.2 M images out of the initial set of 3.1 M. Manual inspection confirms 100\% precision, implying that no false positives are retained.

\textbf{Type Classification and Title Extraction.}
We use GPT-4o to obtain each chart’s type and title. We filter out the charts that do not belong to the nine predefined types, leaving 577k chart images. If an explicit title is not detected, we assign the placeholder “not specified” to serve as a negative instruction, aiming to reduce hallucinations ~\citep{liu2023mitigating}. Manual evaluation shows an accuracy of 99\%, with minor error due to ambiguous title placement within the chart.

\textbf{Extracting Type-Specific Information.}
In accordance with our caption schema, we obtain structural components and key insights. We use GPT-4o for coarse-grained tasks such as identifying overall trends, while using Claude 3.5 Sonnet for more fine-grained tasks (e.g., locating exact max or min values). Preliminary experiments find that GPT-4o struggles to extract precise numerical values. Experiment details for this model selection are provided in Appendix \ref{appendix:task_allocation_experiment}. If no information is extracted, it is labeled “not specified”. Extracted information from the previous and current stages is accumulated in a semi-structured format as shown in Figure \ref{fig:pipeline_figure}. Manual evaluation yields 94\% accuracy, with some misinterpretation  occurring in logarithmic-scale charts, scatter plots with no distinct correlations, and charts containing inset plots.

\textbf{Finalizing the Caption.}
The semi-structured data is transformed into sentence-level captions. Given the relative simplicity of this stage, we use GPT-4o-mini to perform the transformation. Manual evaluation confirms that all transformations are accurate and preserve information.

\subsection{Human Verification via Cycle Consistency}
\begin{figure}[t]
\begin{center}
\includegraphics[width=\linewidth]{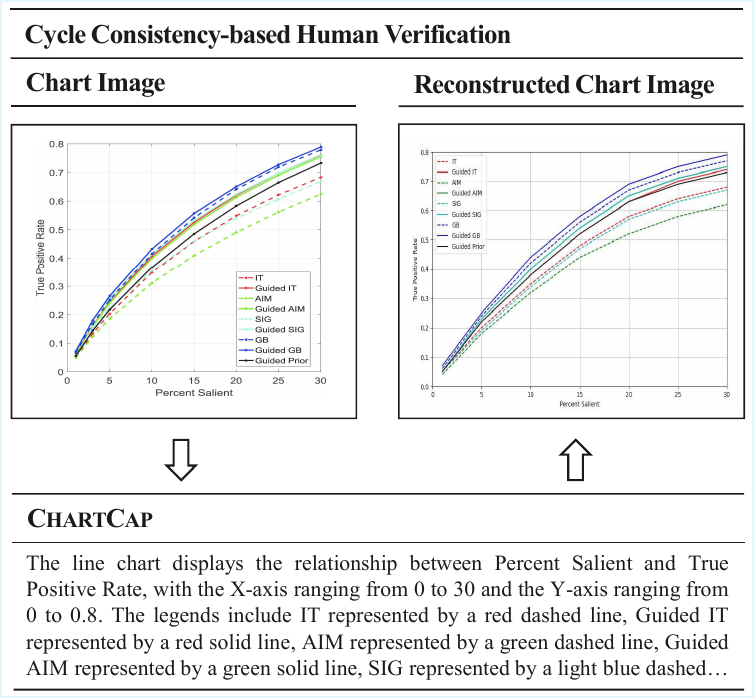}
\end{center}
\vspace{-12pt}
\caption{
An illustration of the cycle consistency-based human verification  for \ChartCap. The original chart image (left) is compared with a reconstructed one (right) using a Python code from the caption (bottom). This process enables efficient human verification by assessing the accuracy and informativeness of the generated captions through visual consistency.
}
\label{fig:cycle_consistency}
\end{figure}

Despite the high performance of proprietary models, human verification remains indispensable for guaranteeing the quality of \ChartCap. However, manually inspecting vast numbers of image-caption pairs is prohibitively time-consuming and expensive. To address this challenge, we introduce a cycle consistency-based human verification process, taking advantage of the millisecond-scale speed of human visual perception \citep{amano2006estimation}, 
as illustrated in Figure~\ref{fig:cycle_consistency}.

We generate Python code using Claude 3.5 Sonnet to recreate chart images from captions and then compare the reconstructed chart images with the originals.  Applying human verification to 68K samples, we finalize a 56K test set. To validate the logical soundness of this verification process, we conduct qualitative and quantitative evaluation, detailed in Appendix \ref{appendix:cycle_consistency_validation}. Our findings are as follows.
\begin{enumerate}
    \item Compared to direct image-caption comparison, our verification process is approximately 24 times faster while maintaining an F1 score of 95\%.
    \item Our verification process ensures both caption correctness and informativeness, making it well-suited for \ChartCap's dense-captioning objectives.
\end{enumerate}

\subsection{The Visual Consistency Score}
How can we evaluate that a generated caption is faithful to its corresponding chart image? 
We believe that the best caption would correctly reproduce the chart, analogous to that a best generative model $P(x)$ is the one that can generate data $x$ themselves.

This intuition leads us to propose a new captioning evaluation metric named the \textit{Visual Consistency Score} (VCS), thanks to recent prominence of LLMs. Unlike natural images, charts have a unique characteristic: they can be deterministically generated from an intermediate modality—namely, code. Leveraging this property, we convert a given caption \(C_i\) into code \(G_i\), subsequently producing a corresponding chart image \(\hat{I}_i\). By measuring the similarity between this generated chart image \(\hat{I}_i\) and the original chart image \(I_i\), the VCS quantitatively evaluates the accuracy and informativeness of the caption \(C_i\).

The VCS is computed by a two-stage procedure, code generation and image comparison. 
Given a caption \(C_i\), an LLM is used to generate Matplotlib code \(G_i\) for recreating the chart. If \(G_i\) fails to execute, the code and runtime error message are supplied back to the LLM for debugging. This process is repeated until code execution succeeds, yielding a valid \(G_i\), which is then executed to generate the chart image \(\hat{I}_i\). The similarity between \(I_i\) and \(\hat{I}_i\) is computed using a cosine similarity with a vision encoder.
Finally, the VCS is the average similarity across all \(N\) samples:
\[
  \text{\textbf{Visual Consistency Score}}
  \;=\;
  \frac{1}{N}
  \sum_{i=1}^{N}
  \text{Sim}(I_i, \hat{I}_i).
\]

\paragraph{OCRScore.}
To evaluate how well textual elements are preserved, Optical Character Recognition (OCR) can be applied to both \(I_i\) and \(\hat{I}_i\). Let \(\mathcal{T}_i\) and \(\hat{\mathcal{T}}_i\) be the sets of text strings extracted from \(I_i\) and \(\hat{I}_i\), respectively. The OCRScore is as an F1 score based on precision (\(P\)) and recall (\(R\)):
\[
\begin{aligned}
P =
  \frac{\sum_{i=1}^{N} \bigl|\mathcal{T}_i \,\cap\, \hat{\mathcal{T}}_i\bigr|}
       {\sum_{i=1}^{N} \bigl|\hat{\mathcal{T}}_i\bigr|},
\quad\quad
R = 
  \frac{\sum_{i=1}^{N} \bigl|\mathcal{T}_i \,\cap\, \hat{\mathcal{T}}_i\bigr|}
       {\sum_{i=1}^{N} \bigl|\mathcal{T}_i\bigr|}
\end{aligned}
\]

\[
\textbf{OCRScore} =
  2 \,\cdot\, \frac{P \times R}{P + R}.
\]

Both VCS and OCRScore exhibit the highest agreement rates with human judgments among automated evaluation metrics such as BERTScore, demonstrating their practicality as reliable, scalable, and effective metrics for evaluating chart-caption quality. Detail of validation experiment is provided in Appendix \ref{appendix:vcs_validation}.

We use Claude 3.5 Sonnet for code generation, due to its superior performance in generating code~\citep{jimenez2023swe}. For the vision encoder, we employ three variants of SigLIP2~\citep{tschannen2025siglip}, each at a resolution of 512, which achieves state-of-the-art performance across a variety of computer vision benchmarks~\citep{deng2009imagenet, lin2014microsoft, plummer2015flickr30k, thapliyal2022crossmodal}. For OCR, we use PaddleOCR~\citep{paddleocr}.

\subsection{Dataset Analysis}
\label{human_eval:dataset}
\textbf{Visual Consistency Score.}
We evaluate the Visual Consistency  Score and OCRScore on 1K samples from each dataset.
The results are presented in Table~\ref{tab:dataset_vcscore}.
\ChartCap achieves the highest scores among all datasets, indicating that its captions are the most accurate to reconstruct the original chart information. The results indirectly reflect two key aspects: informativeness and the exclusion of extraneous information. Caption informativeness can be partially assessed by the average word count as \ChartCap contains the longest captions with 231.1 words on average.

{\renewcommand{\arraystretch}{1.0}
\begin{table}[t!]
\begin{center}
\begin{adjustbox}{width=\columnwidth}
\begin{tabular}{lccccc}
\toprule
\multirow{2}{*}{\textbf{Dataset}}
& \multicolumn{3}{c}{\textbf{Visual Consistency Score}}
& \multirow{2}{*}{\textbf{OCRScore}} & \multirow{2}{*}{\textbf{Word Count}}\\
\cmidrule(lr){2-4}
& \textbf{Large}
& \textbf{So400M}
& \textbf{Base}
& \\
\midrule
ArxivCap
  & 0.7561 & 0.7421 & 0.7999 & 0.1781 & 43.7\\
ChartSumm
  & 0.8940 & 0.9008  & 0.8898  & 0.2635 & 45.4\\
Chart-to-Text 
  & 0.6925 & 0.7089 & 0.7127 & 0.0951 & 62.2\\
SciCap
    & 0.7861 & 0.8015 & 0.8457 & 0.1843 & 34.5\\
\midrule
\rowcolor{lightgray}\textbf{\ChartCap}
  & \textbf{0.8983}  & \textbf{0.9089}  & \textbf{0.9133}  & \textbf{0.5424} & \textbf{231.1} \\
\bottomrule
\end{tabular}
\end{adjustbox}
\caption{Comparison of real-world chart datasets. The terms Large, So400M, and Base indicate three versions of the SigLIP2 encoder~\citep{tschannen2025siglip}: 
SigLIP2-$\{$large, so400m, base$\}$-512.}
\label{tab:dataset_vcscore}
\end{center}
\vspace{-10pt}
\end{table}}

\begin{figure}[t]
\begin{center}
\includegraphics[width=\linewidth]{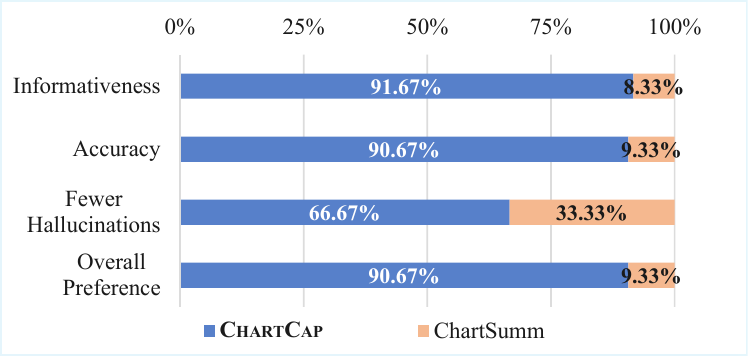}
\end{center}
\vspace{-18pt}
\caption{Results of the head-to-head human evaluation comparing \ChartCap with ChartSumm~\cite{rahman2023chartsumm}.}
\label{fig:human_eval_dataset_comparison}
\end{figure}

\textbf{Human Evaluation.}
We conduct a head-to-head human evaluation by recruiting three annotators via Amazon Mechanical Turk (AMT), comparing 100 samples from \ChartCap and ChartSumm~\cite{rahman2023chartsumm} (the best dataset except ours). Each sample is evaluated based on informativeness, accuracy, fewer hallucinations, and overall preference. Details of human evaluation can be found in Appendix \ref{appendix:human_eval}. As illustrated in Figure \ref{fig:human_eval_dataset_comparison}, \ChartCap consistently outperforms ChartSumm across all evaluated aspects, demonstrating higher overall quality recognized by human.

{\renewcommand{\arraystretch}{1.0}
\begin{table*}[t!]
\begin{center}
\begin{adjustbox}{width=0.90\textwidth}
\begin{tabular}{l c c c c ccc c}
\toprule
\multirow{2}{*}{\textbf{Model}} 
& \multicolumn{4}{c}{\textbf{Reference-based Metrics}} 
& \multicolumn{3}{c}{\textbf{Visual Consistency Score}} 
& \multirow{2}{*}{\textbf{OCRScore}} \\
\cmidrule(lr){2-5} \cmidrule(lr){6-8}
& \textbf{sacreBLEU} & \textbf{ROUGE-L} & \textbf{METEOR} & \textbf{BERTScore}
& \textbf{Large} & \textbf{So400M} & \textbf{Base}
& \\
\midrule
\multicolumn{9}{c}{\textit{\textbf{Proprietary Model}}} \\
\midrule
Claude 3.5 Sonnet 
  & 5.35 & 0.2265 & 0.2131 & 0.6606
  & 0.8834 & 0.8771 & 0.8976 & 0.4868 \\
\midrule
\multicolumn{9}{c}{\textit{\textbf{Chart Expert Models}}} \\
\midrule
ChartGemma-2B
  & 0.73 & 0.1607 & 0.1082 & 0.5946
  & 0.8314 & 0.8184 & 0.8565 & 0.2351 \\
ChartIns-Llama2-7B 
  & 0.62 & 0.1144 & 0.0814 & 0.5157
  & 0.6947 & 0.6759 & 0.7541 & 0.1830 \\
\midrule
\multicolumn{9}{c}{\textit{\textbf{Open-source Models}}} \\
\midrule
InternVL2.5-78B
  & 8.15 & 0.2510 & 0.2336 & 0.6642
  & 0.8841 & 0.8766 & 0.8985 & 0.4677 \\
InternVL2.5-38B
  & 5.88 & 0.2331 & 0.2020 & 0.6551
  & 0.8790 & 0.8700 & 0.8965 & 0.4300 \\
InternVL2.5-26B
  & 5.32 & 0.2350 & 0.1972 & 0.6546
  & 0.8751 & 0.8674 & 0.8873 & 0.4144 \\
InternVL2.5-8B
  & 3.60 & 0.1770 & 0.1577 & 0.6139
  & 0.8485 & 0.8372 & 0.8720 & 0.3456 \\
\rowcolor{lightgray}\textbf{InternVL2.5-8B$_{\text{\ChartCap}}$}
  & \underline{19.47} & \underline{0.3393} & \underline{0.3729} & \underline{0.7238}
  & \underline{0.8913} & \underline{0.8828} & \underline{0.9068} & \underline{0.5089} \\
Phi3.5-Vision-4B
  & 8.41 & 0.2466 & 0.2501 & 0.6626
  & 0.8433 & 0.8323 & 0.8696 & 0.4875 \\
\textbf{Phi3.5-Vision-4B$_{\text{Original}}$}
  & 0.09 & 0.0782 & 0.0384 & 0.5066
  & 0.7782 & 0.7655 & 0.8137 & 0.1438 \\
\textbf{Phi3.5-Vision-4B$_{\text{ChartSumm}}$}
  & 1.31 & 0.1509 & 0.1322 & 0.6008
  & 0.8002 & 0.7873 & 0.8207 & 0.2042 \\
\rowcolor{lightgray}\textbf{Phi3.5-Vision-4B$_{\text{\ChartCap}}$}
  & \textbf{23.82} & \textbf{0.3900} & \textbf{0.4084} & \textbf{0.7427}
  & \textbf{0.8933} & \textbf{0.8829} & \textbf{0.9092} & \textbf{0.5179} \\
\bottomrule
\end{tabular}
\end{adjustbox}
\caption{Results of reference-based metrics, Visual Consistency Scores, and OCRScore on the \ChartCap test set.}
\label{tab:chartcap_automatic_metrics_combined}
\end{center}
\vspace{-10pt}
\end{table*}}

\section{Experiments}

We demonstrate the effectiveness of our \ChartCap dataset: first, 
we show that VLMs fined-tuned on \ChartCap attain strong dense captioning performance in terms of reference-based metrics, human evaluation, and the Visual Consistency Score. 
Second, we present that \ChartCap-trained captioning models show compelling zero-shot captioning on two human-annotated benchmarks, VisText~\citep{tang2023vistext} and Chart-to-Text~\citep{kantharaj2022chart}.

\textbf{Base Models.}
We experiment with open-source, chart expert, and proprietary captioning models.
For open-source models, we use InternVL2.5-78B~\citep{chen2024expanding}, InternVL2.5-38B, InternVL2.5-26B, InternVL2.5-8B, and Phi3.5-vision-4B~\citep{abdin2024phi}. For chart expert models, we employ ChartGemma-2B~\citep{masry2024chartgemma} and ChartInstruct-Llama2-7B~\citep{masry2024chartinstruct}. For proprietary models, we use the Claude 3.5 Sonnet~\citep{claude3.5}, which not only achieves the best performance in our dataset but also reports the state-of-the-art performance on ChartQA~\citep{masry2022chartqa} and CharXiv~\citep{wang2024charxiv}. Additional baselines are provided in Appendix~\ref{appendix:baselines}.

\textbf{Experiment Setup and Metrics.}
All models are prompted with the same instruction: \textit{"Please provide a detailed caption for the chart."} along with the chart image as input. For  metrics, we use SacreBLEU~\citep{post2018call}, ROUGE~\citep{lin2004rouge}, METEOR~\citep{banerjee2005meteor}, and BERTScore~\citep{zhang2019bertscore}, with our Visual Consistency Score and OCRScore.

\textbf{Training Settings.}
We perform supervised fine-tuning using LoRA fine-tuning~\citep{hu2021lora} on InternVL2.5-8B and Phi3.5-vision-4B on the \ChartCap training set (509K). We also fine-tune Phi3.5-vision-4B using 250K original captions from ArxivCap, ChartSumm-Knoema, and ChartCheck. Additionally, We fine-tune Phi-3.5-Vision-4B on the entire training set of ChartSumm. Fine-tuned models are denoted with the name of the training dataset (e.g., Phi3.5-Vision-4B$_{\text{\ChartCap}}$).

\begin{figure}[t]
\includegraphics[width=\columnwidth]{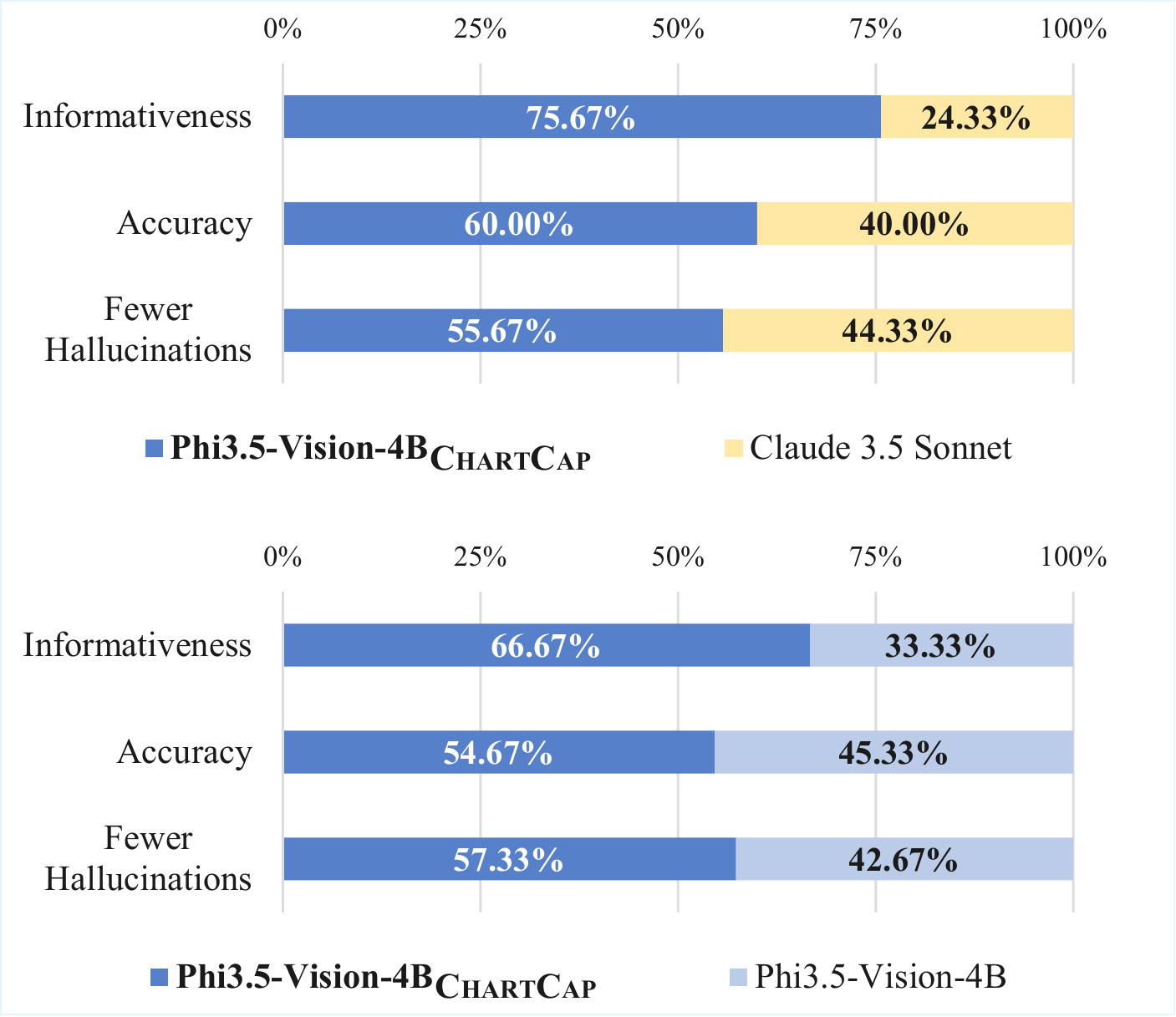}  
\vspace{-18pt}
\caption{Results of human evaluation results comparing Phi3.5-Vision-4B$_{\text{\ChartCap}}$ against Claude 3.5 Sonnet (top) and Phi3.5-Vision-4B (bottom) on the \ChartCap test set.}
\label{fig:human_eval_chartcap}
\end{figure}

\subsection{Results on the \textbf{\ChartCap}}
\textbf{Reference-based Metrics.}
Table \ref{tab:chartcap_automatic_metrics_combined} presents the results of the reference-based metrics on the \ChartCap test set. Both InternVL2.5-8B$_{\text{\ChartCap}}$ and Phi3.5-Vision-4B$_{\text{\ChartCap}}$ achieve higher scores than all baseline models across all evaluation metrics. In contrast, Phi3.5-Vision-4B$_{\text{Original}}$ and Phi3.5-Vision-4B$_{\text{ChartSumm}}$ records significantly lower scores, even shows degradation of its base model. These results indicate that our fine-tuned models generate captions that align closely with the human-verified reference captions of \ChartCap, which accurately capture the structural components and key insights of the charts sufficiently.

\textbf{Human Evaluation.}
\label{human_evaluation:chartcap}
We conduct a human evaluation to assess caption accuracy, informativeness, and the extent of hallucination, as reference-based metrics do not measure absolute caption quality and struggle to effectively assess the degree of hallucination~\citep{kryściński2019evaluatingfactualconsistencyabstractive, maynez-etal-2020-faithfulness}. For human evaluation, we select Phi3.5-Vision-4B$_{\text{\ChartCap}}$ that shows the highest score on the reference-based metrics on \ChartCap and compare head-to-head with one proprietary model (Claude 3.5 Sonnet) and one open-source model (Phi3.5-Vision-4B). We randomly sample 100 captions generated by each model and recruit three crowd workers via AMT to select the better caption based on three aspects: (1) informativeness, (2) accuracy, and (3) fewer hallucinations. Further details on the human evaluation are provided in the Appendix \ref{appendix:human_eval}.

As shown in Figure~\ref{fig:human_eval_chartcap}, Phi3.5-Vision-4B$_{\text{\ChartCap}}$ ranks consistently higher in all three human evaluation criteria. This consistency suggests that human judges feel Phi3.5-Vision-4B$_{\text{\ChartCap}}$ generates more informative and accurate captions with fewer hallucinations compared to baseline models. Notably, despite having a smaller model size, Phi3.5-Vision-4B$_{\text{\ChartCap}}$ surpasses the strong proprietary model, Claude 3.5 Sonnet, according to human judgments. This highlights fine‐tuning on high‐quality data could overshadow the model scale.

\textbf{The Visual Consistency Score.}
Table~\ref{tab:chartcap_automatic_metrics_combined} also presents the Visual Consistency Score and OCRScore for the \ChartCap test set. Both InternVL2.5-8B$_{\text{\ChartCap}}$ and Phi3.5-Vision-4B$_{\text{\ChartCap}}$ exhibit higher Visual Consistency Score and OCRScore relative to all baselines, signifying that the captions they produce align more closely with the ground‐truth chart structure and text elements. This stronger grounding in chart content further explains why they offer more accurate, informative, and low‐hallucination captions than non‐fine‐tuned variants or other baselines.

\subsection{Results on Other Human-Verified 
Benchmarks}
We evaluate the zero-shot performance of previous captioning models on other human-verified benchmarks. We first test  on the entire VisText test set, consisting of synthetic charts with human-authored captions following the caption schema from \citep{lundgard2021accessible}. We also evaluate the models on the 1K PEW subset of Chart-to-Text, a real-world dataset whose subset has undergone human verification. The experiment setup is the same as the previous experiment.

\textbf{Human Evaluation.}
\label{human_evaluation:vistext}
We conduct a human evaluation on 100 samples from the VisText test set, comparing captions generated by Phi3.5-Vision-4B$_{\text{\ChartCap}}$ with Claude 3.5 Sonnet and the human-authored ground-truth captions, under the same evaluation protocol.

\begin{figure}
\vspace{-12pt}
    \centering
    \includegraphics[width=\columnwidth]{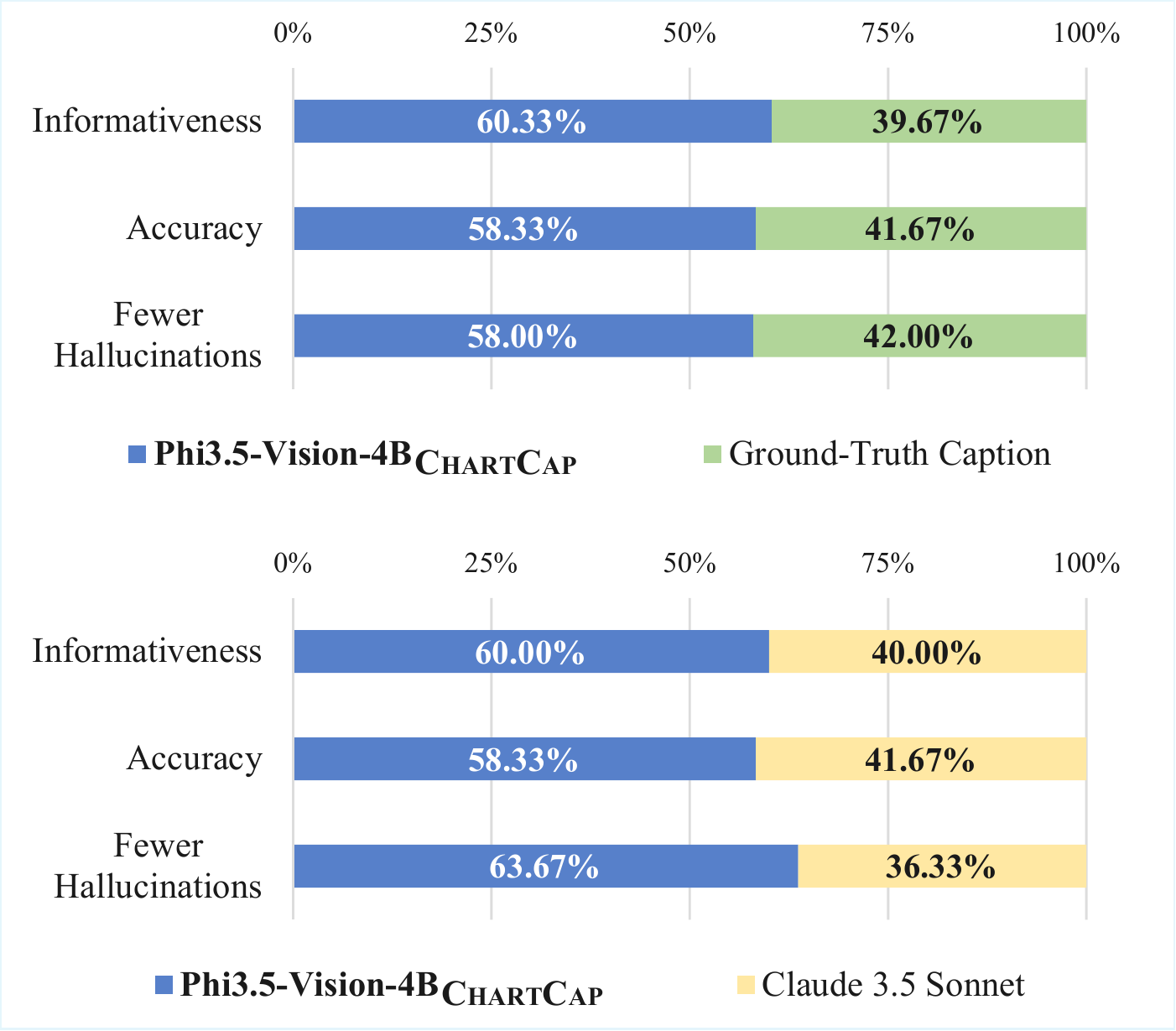}
    \vspace{-12pt}
    \caption{Human evaluation results comparing Phi3.5-Vision-4B$_{\text{\ChartCap}}$ against ground-truth captions (top) and Claude 3.5 Sonnet (bottom) on the VisText test set.}
    \label{fig:humaneval_vistext}
\end{figure}

As shown in Figure \ref{fig:humaneval_vistext}, Phi3.5-Vision-4B$_{\text{\ChartCap}}$ outperforms both the ground-truth captions and Claude 3.5 Sonnet across all three evaluation aspects. Interestingly, human annotators judge that the model fine-tuned on \ChartCap can generate better chart descriptions than human-authored ground-truth captions across
all axes by a large margin.

\textbf{The Visual Consistency Score.}
Table~\ref{tab:vistext_vcscore} --~\ref{tab:c2t_vcscore} present the Visual Consistency Score for the VisText test set, and the PEW subset of the Chart-to-Text dataset, respectively. As shown in both tables, InternVL2.5-8B$_{\text{\ChartCap}}$ and Phi3.5-Vision-4B$_{\text{\ChartCap}}$ achieve the highest Visual Consistency Scores and competitive OCRScores among all baseline models. Again, these two models surpass even the human-annotated ground-truth captions in accurately reconstructing the original chart images. In particular, for the VisText dataset, only the \ChartCap-trained models outperform the human-authored ground-truth captions.
The results also highlight the generalizability and effectiveness of captioning models trained with \ChartCap.

{\renewcommand{\arraystretch}{1.0}
\begin{table}[t!]
\begin{adjustbox}{width=\columnwidth}
\begin{tabular}{lccc}
\toprule
\multirow{2}{*}{\textbf{Model}}
& \multicolumn{2}{c}{\textbf{Visual Consistency Score}}
& \multirow{2}{*}{\textbf{OCRScore}} \\
\cmidrule(lr){2-3}
& \textbf{Large}
& \textbf{So400M}
& \\
\midrule
\textbf{Ground-truth Caption}
    & 0.9172 & 0.9151 & 0.3407 \\
\midrule
Claude 3.5 Sonnet
  & 0.8970 & 0.9008 & 0.3286 \\
InternVL2.5-8B
  & 0.9093 & 0.9082 & 0.3172 \\
\rowcolor{lightgray}\textbf{InternVL2.5-8B$_{\text{\ChartCap}}$}
  & \underline{0.9401} & \underline{0.9355} & 0.3360 \\
Phi3.5-Vision-4B 
  & 0.8809 & 0.8814 & \textbf{0.3826} \\
\rowcolor{lightgray}\textbf{Phi3.5-Vision-4B$_{\text{\ChartCap}}$}
  & \textbf{0.9443}  & \textbf{0.9382}  & \underline{0.3414} \\
\bottomrule
\end{tabular}
\end{adjustbox}
\caption{Visual Consistency Scores and OCRScore on the VisText test set.}
\label{tab:vistext_vcscore}
\end{table}}

\begin{figure}
    \centering
    \vspace{-5pt}
    \includegraphics[width=\columnwidth]{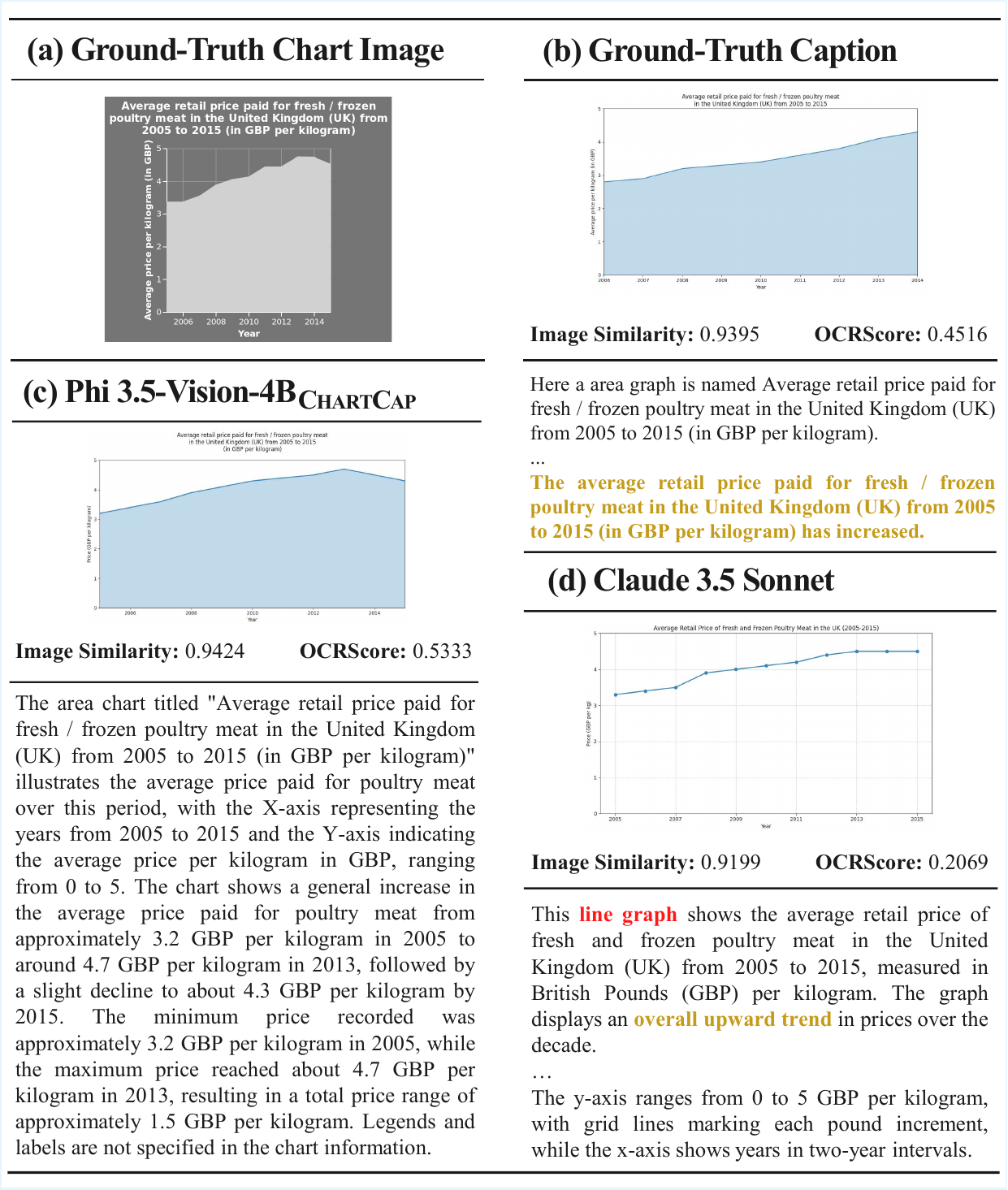}
    \caption{Qualitative examples from VisText, comparing (a) the ground-truth chart image with captions and their reconstructed charts from the captions of (b) human-authored ground-truth, (c) Phi3.5-Vision-4B$_{\text{\ChartCap}}$, and (d) Claude 3.5 Sonnet.}
    \vspace{1pt}
    \label{fig:qual_result}
\end{figure}

\textbf{Qualitative Examples.}
Figure~\ref{fig:qual_result} compares the captions and their reconstructed charts generated by Phi3.5-Vision-4B$_{\text{\ChartCap}}$, human-authored ground-truth caption, and Claude 3.5 Sonnet for a chart from VisText. The caption generated by Phi3.5-Vision-4B$_{\text{\ChartCap}}$ provide precise and detailed descriptions of both the chart’s structural components and data. As a result, its reconstructed chart closely resembles the original ground-truth, and consequently achieve the highest VCS and OCRScore. In contrast, the human-authored caption describes data trends in a simplified manner (e.g., merely stating that values increase), resulting in reconstructed charts that exhibit overly simplified data trends. Similarly, the caption generated by Claude 3.5 Sonnet describes the data trend without sufficient detail and incorrectly classifies the chart type, leading to a reconstructed chart that not only simplifies the trend but also displays an incorrect chart type.

{\renewcommand{\arraystretch}{1.0}
\begin{table}[t!]
\begin{adjustbox}{width=\columnwidth}
\begin{tabular}{lccc}
\toprule
\multirow{2}{*}{\textbf{Model}}
& \multicolumn{2}{c}{\textbf{Visual Consistency Score}}
& \multirow{2}{*}{\textbf{OCRScore}} \\
\cmidrule(lr){2-3}
& \textbf{Large}
& \textbf{So400M}
& \\
\midrule
\textbf{Ground-truth Caption}
    & 0.6925 & 0.7089 & 0.0951 \\
\midrule
Claude 3.5 Sonnet
  & 0.7495 & 0.7616 & 0.1603 \\
InternVL2.5-8B
  & 0.7362 & 0.7478 & 0.1272 \\
\rowcolor{lightgray}\textbf{InternVL2.5-8B$_{\text{\ChartCap}}$}
  & \underline{0.7946} & \underline{0.8013} & \textbf{0.1833} \\
Phi3.5-Vision-4B 
  & 0.7370 & 0.7490 & 0.1786 \\
\rowcolor{lightgray}\textbf{Phi3.5-Vision-4B$_{\text{\ChartCap}}$}
  & \textbf{0.7999}  & \textbf{0.8075} & \underline{0.1789} \\
\bottomrule
\end{tabular}
\end{adjustbox}
\caption{Visual Consistency Scores and OCRScore on the PEW subset of the Chart-to-Text.}
\label{tab:c2t_vcscore}
\end{table}}

\section{Conclusion}
We introduced \ChartCap, a large-scale dataset of 565K real-world chart images paired with type-specific captions that include both structural components and key insights in a dense manner while minimizing extraneous information. We constructed \ChartCap via a four-phase caption generation pipeline with systematically devised caption-schema and cycle consistency-based human verification. We also proposed the Visual Consistency Score to assess caption quality by measuring the consistency between the original charts and the ones generated from captions. Models fine-tuned on \ChartCap substantially enhance the quality of chart captions, even generates better caption than strong proprietary baseline and human annotated captions. 

As a limitation, \ChartCap utilizes a caption-schema built upon the blueprint of nine chart types defined by VLAT~\citep{lee2016vlat}, which restricts the diversity of chart types covered. It is an interesting future work to expand the captioning schema and integrate it into the proposed pipeline, which could enable the creation of a more diverse large-scale dataset.

\section*{Acknowledgments}
We thank the anonymous reviewers and Chaeyoung Lim for their valuable comments. This work is supported by the Samsung Electronics' University R\&D program [Efficient fine-tuning of large multimodal models for domain-specific figure description], Institute of Information \& Communications Technology Planning \& Evaluation (IITP) grant funded by the Korea government (MSIT) (No.~RS-2019-II191082, SW StarLab, No.~RS-2022-II220156, Fundamental research on continual meta-learning for quality enhancement of casual videos and their 3D metaverse transformation, and No.~RS-2021-II211343, Artificial Intelligence Graduate School Program of Seoul National University), National Research Foundation of Korea (NRF) grant funded by the Korea government (MSIT) (No.~2023R1A2C2005573), and Seoul R\&BD Program (VC230004) through the Seoul Business Agency (SBA) funded by The Seoul Metropolitan Government. Gunhee Kim is the corresponding author.

{
    \small
    \bibliographystyle{ieeenat_fullname}
    \bibliography{main}
}


\appendix

\clearpage
\setcounter{page}{1}
\maketitlesupplementary

\section{Type-specific Caption Schema}
\label{appendix:type-specific_caption_schema}

\subsection{Caption Schema for Structural Description}
The caption schema that defines the structural elements included in each chart type is shown in Table~\ref{tab:caption_schema_structural_description}.

{\renewcommand{\arraystretch}{1.0}
\begin{table}[t!]
\centering
\begin{adjustbox}{width=0.48\textwidth}
\begin{tabular}{lcccccc}
    \toprule
    \makecell[l]{\textbf{Chart Type}} & \makecell{\textbf{Title}} & \makecell{\textbf{Axes}} & \makecell{\textbf{Categories}} & \makecell{\textbf{Bubble}} & \makecell{\textbf{Legends}} & \makecell{\textbf{Labels}} \\
    \midrule
    \makecell[l]{Line} & \cmark & \cmark &  &  & \cmark & \cmark \\
    \makecell[l]{Bar} & \cmark & \cmark & \cmark &  & \cmark & \cmark \\
    \makecell[l]{Pie} & \cmark &  & \cmark &  & \cmark & \cmark \\
    \makecell[l]{Histogram} & \cmark & \cmark &  &  & \cmark & \cmark \\
    \makecell[l]{Scatter} & \cmark & \cmark &  &  & \cmark & \cmark \\
    \makecell[l]{Area} & \cmark & \cmark &  &  & \cmark & \cmark \\
    \makecell[l]{Bubble} & \cmark & \cmark & \cmark & \cmark & \cmark & \cmark \\
    \bottomrule
\end{tabular}

\end{adjustbox}
\caption{Caption schema specifying the structural elements required for each chart type. For readability, choropleth maps and treemaps are excluded due to their distinct characteristics. Choropleth maps include title, base map, color scale, geographic labels, data classes, and north arrow. Treemaps include title, tiles, hierarchy levels, and color coding.}
\label{tab:caption_schema_structural_description}
\end{table}}

\subsection{Caption Schema for Key Insights}
The caption schema that specifies the key insights to be included for each chart type can be found in Table~\ref{tab:caption_schema_key_insights}. In the case of ``Retrieve Value'', the task involves reading data points to answer a question, making it inapplicable to the captioning task. Therefore, if the data points had labels, those were extracted; otherwise, the initial, middle, and final data values were extracted in the caption.

{\renewcommand{\arraystretch}{1.0}
\begin{table*}[t!] 
\begin{center}
\begin{adjustbox}{width=1.0\textwidth}
\begin{tabular}{lcccccccc}
    \toprule
    \makecell[l]{\textbf{Chart Type}} & \makecell{\textbf{Retrieve}\\\textbf{Value}} & \makecell{\textbf{Find}\\\textbf{Extremum}} & \makecell{\textbf{Make}\\\textbf{Comparison}} & \makecell{\textbf{Determine}\\\textbf{Range}} & \makecell{\textbf{Find}\\\textbf{Correlations /}\\\textbf{Trend}} & \makecell{\textbf{Characterize}\\\textbf{Distribution}} & \makecell{\textbf{Find}\\\textbf{Clusters}} & \makecell{\textbf{Find}\\\textbf{Anomalies}} \\
    \midrule
    \makecell[l]{Line} & \cmark & \cmark & \cmark & \cmark & \cmark &  &  &  \\
    \makecell[l]{Bar} & \cmark & \cmark & \cmark & \cmark &  &  &  &  \\
    \makecell[l]{Pie} & \cmark & \cmark & \cmark &  &  &  &  &  \\
    \makecell[l]{Histogram} & \cmark & \cmark & \cmark &  &  & \cmark &  &  \\
    \makecell[l]{Scatter} & \cmark & \cmark & \cmark & \cmark & \cmark & \cmark & \cmark & \cmark \\
    \makecell[l]{Area} & \cmark & \cmark & \cmark & \cmark & \cmark &  &  &  \\
    \makecell[l]{Bubble} & \cmark & \cmark & \cmark & \cmark & \cmark & \cmark & \cmark & \cmark \\
    \makecell[l]{Choropleth Map} & \cmark & \cmark & \cmark &  &  &  &  &  \\
    \makecell[l]{Treemap} & \cmark & \cmark & \cmark &  &  &  &  &  \\
    \bottomrule
\end{tabular}
\end{adjustbox}
\caption{Caption schema specifying the key insights required for each chart type.}
\label{tab:caption_schema_key_insights}
\end{center}
\end{table*}}

\section{Prompt Demonstrations}
We present the prompts used in the dataset-generation pipeline and in chart regeneration for the cycle-consistency-based human-verification process and the Visual Consistency Score.

\begin{itemize}
    \item \textbf{Filtering Non-Chart Images}: See Table~\ref{tab:prompt_stage_1}
    \item \textbf{Type Classification and Title Extraction}: See Table~\ref{tab:prompt_stage_2}
    \item \textbf{Retrieving Type-Specific Information}: See Tables~\ref{tab:prompt_stage_3_gpt} and~\ref{tab:prompt_stage_3_claude}
    \item \textbf{Finalizing the Caption}: See Table~\ref{tab:prompt_stage_4}
    \item \textbf{Chart Regeneration}: See Table~\ref{tab:prompt_cycle_consistency}
    \item \textbf{Code Debugging}: See Table~\ref{tab:code_debugging}
\end{itemize}

\begin{table*}[htbp]
\scriptsize
\centering
\begin{tabular}{@{}p{\linewidth}@{}}
\toprule
\textbf{Filtering non-chart images.}\\
\midrule
Please determine whether the image contains a single, data-driven chart only. A data-driven chart is a visual representation directly based on numerical data. Note that an inset chart (a smaller chart embedded within a larger chart) is not considered a multi-chart. \\
\hspace{2em}- If the image consists exclusively of a single data-driven chart (with no additional visuals, such as natural images, illustrations, conceptual diagrams, or schematics) and does not contain multiple subplots, respond with: \\
\hspace{4em}Single-Chart: yes \\
\\
\hspace{2em}- If the image contains any non-data-driven elements (e.g., natural images, illustrations, conceptual diagrams, schematics) or features multiple charts/subplots, respond with: \\
\hspace{4em}Single-Chart: no \\
\\
Follow the exact response format:\\
Single-Chart: \\
\bottomrule
\end{tabular}
    \caption{Prompt used for filtering non-chart images.}
    \label{tab:prompt_stage_1}
\end{table*}

\begin{table*}[htbp]
\scriptsize
\centering
\begin{tabular}{@{}p{\linewidth}@{}}
\toprule
\textbf{Type Classification and Title Extraction.}\\
\midrule
\textbf{[System]}\\
You are an expert in data visualization and chart interpretation. Your task is to provide accurate analysis of charts such as identifying and classifying the chart.\\
\\
\textbf{[User]}\\
Please analyze the image to classify the chart type(s) and extract the main title according to the instructions below.\\
\hspace{2em}- Identify the chart type(s) from the following list: [line, bar, pie, histogram, scatter, area, bubble, choropleth map, treemap].\\
\hspace{2em}- If it belongs to multiple chart types, list them separated by commas (e.g., ``bar'', ``line''). If it does not match any listed chart types or is a 3D visualization, respond with:\\
\hspace{2em}Type: none\\
\hspace{2em}Title: not specified\\
\\
\hspace{2em}- If the image contains one or more valid chart types, extract the main title of the chart. If the title is not visible or unclear, respond with 'not specified'.\\
\\
Follow the exact response format:\\
\\
Type: $<$list of chart\_type(s) or 'none'$>$\\
Title: $<$chart\_title or 'not specified'$>$\\
\bottomrule
\end{tabular}
\caption{Prompt used for type classification and title extraction.}
\label{tab:prompt_stage_2}
\end{table*}

\begin{table*}[htbp]
\scriptsize
\centering
\begin{tabular}{@{}p{\linewidth}@{}}
\toprule
\textbf{Extracting Type-specific Information (Coarse-grained)}\\
\midrule
\textbf{[System]}\\
You are an expert in data visualization and chart interpretation. Your task is to provide accurate analysis of charts such as identifying the components, key trends, and insights, without making any guesses.\\
\\
\textbf{[User]}\\
Identify and describe the components of the given line chart. Only explain the components if they exist; otherwise, respond with not specified. Do not guess or include information not visible in the image, except for approximations in axes ranges, retrieving value of data points, and determining data point ranges.\\
If the chart is multi-series, grouped, or includes an inset chart, compute and report information for each data series or category separately.\\
\\
* Type: Provide the type or types of the chart from line chart, bar chart, pie chart, histogram, scatter plot, area chart, bubble chart, choropleth map, and treemap.\\
* Legends: Identify any legends or keys that globally explain symbols, colors, or data series.\\
* Labels: Identify specific labels that annotate or describe individual elements, such as data points, bars, or segments of a chart. Exclude axis labels and legends.\\
* Data Comparison: Highlight specific similarities or differences between data points or categories in the chart. Focus on relative comparisons rather than extracting or explaining precise values. Avoid analyzing overall trends.\\
* Data Correlations/Trends: Analyze patterns or relationships between variables, noting any trends.\\
\\
Only respond with the analyzed results, avoiding any additional statements or extraneous text.\\
Follow the exact response format:\\
$<$attribute 1$>$: $<$analysis result$>$\\
$<$attribute 2$>$: $<$analysis result$>$\\
...\\
\bottomrule
\end{tabular}
    \caption{Prompt used for extracting coarse-grained, type-specific information from line charts.}
    \label{tab:prompt_stage_3_gpt}
\end{table*}

\begin{table*}[htbp]
\scriptsize
\centering
\begin{tabular}{@{}p{\linewidth}@{}}
\toprule
\textbf{Extracting Type-specific Information (Fine-grained).}\\
\midrule
\textbf{[System]}\\
You are an expert in data visualization and chart interpretation. Your task is to provide accurate analysis of charts such as identifying the components, retrieving data points, statistics, key trends, insights, without any guesses.\\
\\
\textbf{[User]}\\
Identify and describe the components of the given line chart. Only explain the components if they exist; otherwise, respond with not specified. Do not guess or include information not visible in the image, except for approximations in axes ranges, retrieving value of data points, and determining data point ranges.\\
If the chart is multi-series, grouped, or includes an inset chart, compute and report information for each data series or category separately.\\
\\
* Axes: Describe the axes, including titles, units, scales, and ranges. If categories are involved in the axes, list their names as well.\\
* Retrieve Value: Retrieve the coordinates of the initial, middle, and end data points. Additionally, if specific numbers or values are labeled for any data points, also provide the coordinates of those points as well.\\
* Find Extremum: Find the coordinate of the minimum and maximum data points for each data series.\\
* Determine Range: Specify the range (span) of the dependent (response) variable's values from the data points, not the range of the axis.\\
\\
Only respond with the analyzed results, avoiding any additional statements or extraneous text.\\
Follow the exact response format:\\
$<$attribute 1$>$: $<$analysis result$>$\\
$<$attribute 2$>$: $<$analysis result$>$\\
...\\
\bottomrule
\end{tabular}
    \caption{Prompt used for extracting fine-grained, type-specific information from line charts.}
    \label{tab:prompt_stage_3_claude}
\end{table*}

\begin{table*}[htbp]
\scriptsize
\centering
\begin{tabular}{@{}p{\linewidth}@{}}
\toprule
\textbf{Finalizing the Caption.}\\
\midrule
\textbf{[System]}\\
You are an expert in converting provided bullet points into continuous sentences without omitting or adding any information.\\
\\
\textbf{[User]}\\
Generate a natural language caption for the chart based strictly on the provided information. Ensure the caption includes all the details given in the input without omitting anything or adding new information beyond what is explicitly stated.\\
Explicitly mention that information is not provided if it is stated as not provided in the chart information.\\
\\
\text{[Chart Information]}\\
\\
\texttt{\{chart\_info\}}\\
\\
Respond only with the generated caption, including all the information provided.\\
Caption:\\
\bottomrule
\end{tabular}
    \caption{Prompt used for finalizing the caption.}
    \label{tab:prompt_stage_4}
\end{table*}

\begin{table*}[htbp]
\scriptsize
\centering
\begin{tabular}{@{}p{\linewidth}@{}}
\toprule
\textbf{Regenerating a Chart from a Caption.}\\
\midrule
\textbf{[System]}\\
You are an expert in Python and the Matplotlib library. Your task is to generate a complete Python script that precisely reflects every detail in the given chart description, without making any guesses.\\
\\
\textbf{[User]}\\
Generate accurate Python code using Matplotlib library strictly based on the given description about a chart.\\
If the description lacks details about required chart components or data points, omit them from the code instead of making assumptions, but ensure that every detail in the description is included.\\
Instead of using numpy's sin, cos, or exp function, manually define data points to represent the chart if needed.\\
Labels are elements that display and specify data points in the chart. They are different from axis labels (titles).\\
\\
\text{[Description]}\\
``\texttt{\{caption\}}''\\
\\
Respond only the generated code.\\
Code:\\
\bottomrule
\end{tabular}
    \caption{Prompt used for regenerating a chart from a caption.}
    \label{tab:prompt_cycle_consistency}
\end{table*}

\begin{table*}[t]
\scriptsize
\centering
\begin{tabular}{@{}p{\linewidth}@{}}
\toprule
\textbf{Debugging Erroneous Code.}\\
\midrule
\textbf{[System]}\\
You are an expert in Python and the Matplotlib library. Your task is to fix the code based on the provided error message.\\
\\
\textbf{[User]}\\
\text{[Erroneous Code]}\\
``\texttt{\{code\}}''\\
\\
\text{[Error Message]}\\
``\texttt{\{error\_message\}}''\\
\\
Analyze the provided error message and fix the code accordingly. Make only the necessary changes to resolve the error while keeping all correctly functioning attributes unchanged. Return only the corrected code without any explanations or additional output.\\
Corrected Code:\\
\bottomrule
\end{tabular}
    \caption{Prompt used for debugging erroneous code.}
    \label{tab:code_debugging}
\end{table*}

\section{Task Allocation Experiment}
\label{appendix:task_allocation_experiment}
\subsection{Experiment Setup}
This experiment was designed to optimize the effective utilization of GPT-4o and Claude 3.5 Sonnet in extracting accurate information based on the caption schema. The tasks defined in Tables~\ref{tab:caption_schema_structural_description} and~\ref{tab:caption_schema_key_insights} were categorized into fine-grained and coarse-grained tasks, and the performance of each model was evaluated accordingly.

Coarse-grained tasks require broad attention across an image, such as understanding of overall data trends or comparisons between data series: ``Type Classification'', ``Title'', ``Category'', ``Bubble'', ``Legend'', ``Label'', ``Make Comparisons'', ``Find Correlations / Trends'', and ``Characterize Distribution''. In contrast, fine-grained tasks require more localized attention, such as extracting precise numerical values or reading specific data points: ``Axes'', ``Retrieve Value'', ``Find Extremum'', ``Determine Range'', ``Find Clusters'', and ``Find Anomalies''. The experiment was conducted on 100 randomly sampled data, and the performance of both models was manually evaluated for each task category. The prompts used in this experiment can be found in Tables~\ref{tab:prompt_stage_3_gpt} and~\ref{tab:prompt_stage_3_claude}.

\begin{figure}[t]
\includegraphics[width=\columnwidth]{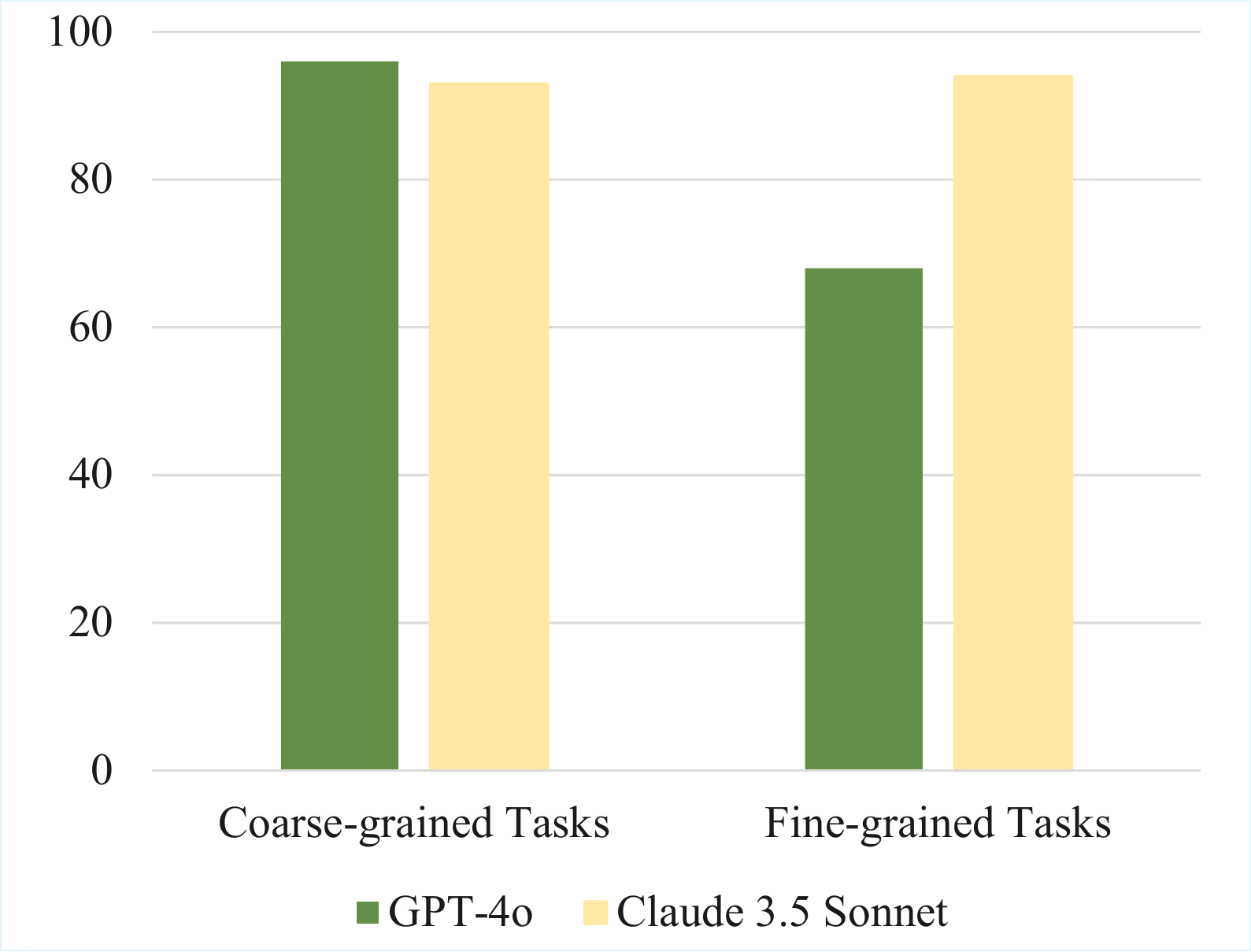}  
\caption{Accuracy of GPT-4o and Claude 3.5 Sonnet on coarse-grained tasks and fine-grained tasks.}
\label{fig:exp_task_allocation}
\end{figure}

\subsection{Results}
The experimental results are shown in Figure~\ref{fig:exp_task_allocation}. For coarse-grained tasks, GPT-4o achieved an accuracy of 96\%, while Claude 3.5 Sonnet achieved 93\%, with GPT-4o demonstrating a slight advantage. The primary sources of error for Claude 3.5 Sonnet were in type classification and title generation, where the model tended to introduce hallucinated information by attempting to predict the overall theme of a chart. Based on these findings, GPT-4o was selected for coarse-grained tasks in our pipeline.

For fine-grained tasks, GPT-4o achieved an accuracy of 68\%, whereas Claude 3.5 Sonnet significantly outperformed it with an accuracy of 94\%. The primary weakness of GPT-4o was its difficulty in accurately reading data coordinates, a critical skill for tasks such as ``Retrieve Value'' and ``Find Extremum'', resulting in incorrect values for maxima, minima, and other key numerical indicators. Because of this fundamental limitation, Claude 3.5 Sonnet was selected for fine-grained tasks in our pipeline.

\begin{figure*}[t]
\begin{center}
\includegraphics[width=0.95\textwidth]{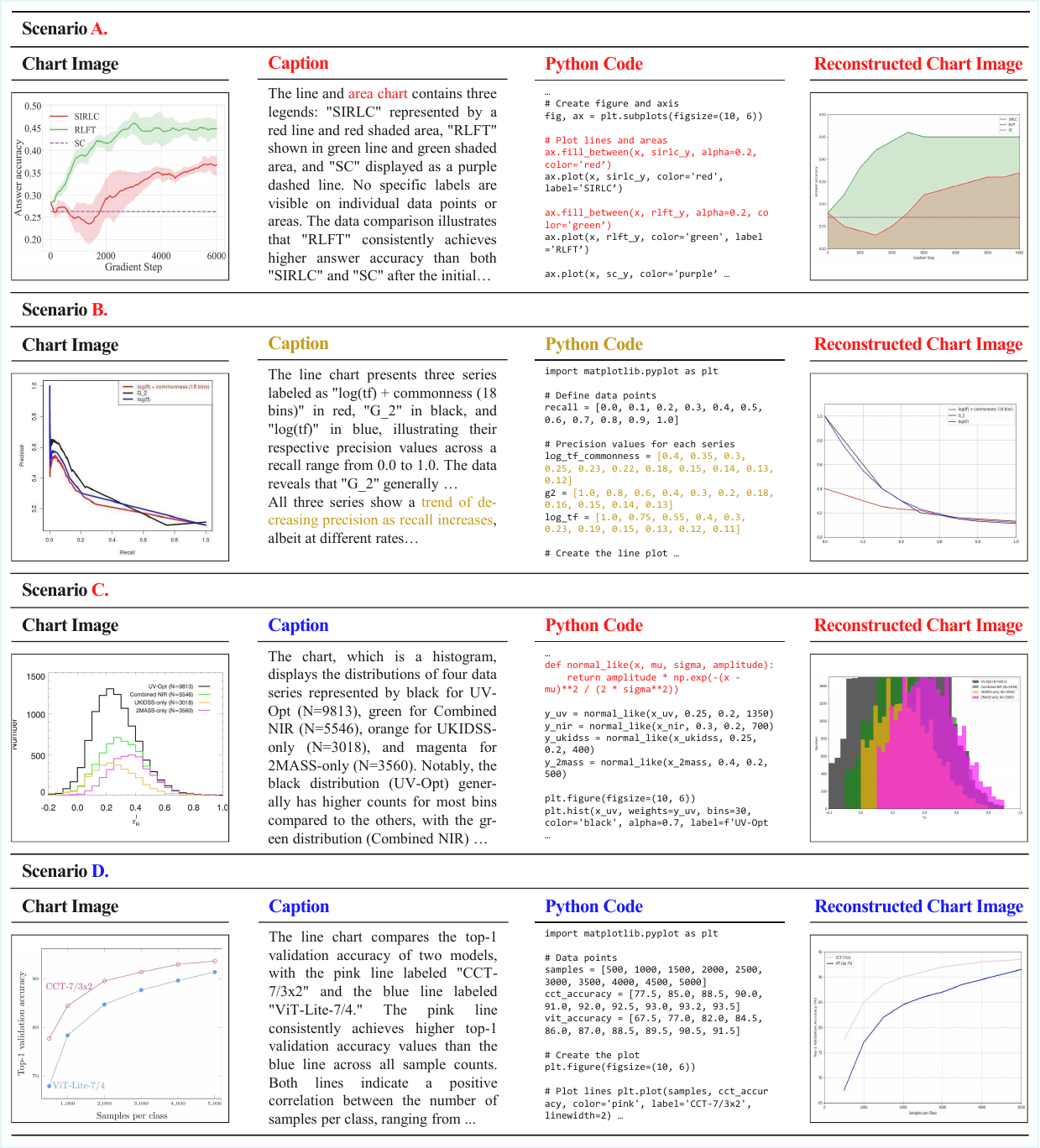}  
\end{center}
\caption{Examples of the four main scenarios that arise during the cycle consistency-based human verification process. In Scenario A, the caption incorrectly describes a \textit{line chart with a shaded area} as an \textit{area chart}. Consequently, the generated code reflects this incorrect information, leading to the reconstruction of a chart that does not match the original. In Scenario B, the caption oversimplifies the data trend by merely describing it as \textit{decreasing}. As a result, the reconstructed chart follows a simple downward trend, failing to capture the original complexity. In Scenario C, an error occurs during the code generation process, leading to the creation of an incorrect chart. Such coding errors were primarily observed when using NumPy's nonlinear functions. Scenario D shows that both the caption and the generated code must be accurate for the reconstructed chart to correctly match the original chart, demonstrating the necessity of precise and informative captions.}
\label{fig:cc_scenarios}
\end{figure*}

\section{Validation of Cycle Consistency-based Human Verification Process}
\label{appendix:cycle_consistency_validation}
\subsection{Quantitative Evaluation}
To quantify the effectiveness of our verification process, we benchmark it against direct chart–caption comparison in terms of both accuracy and efficiency.

\begin{itemize}
    \item \textbf{Accuracy}: On 100 randomly sampled pairs from \ChartCap, our process achieved an F1 score of 94.7\%, with a recall of 90.0\% and precision of 100.0\%,  ensuring that no incorrect captions were falsely validated.
    \item \textbf{Efficiency}: Direct chart-caption comparison required approximately 145 seconds per sample, whereas our process took only 6 seconds per sample, achieving a 24$\times$ speedup.
\end{itemize}

\subsection{Qualitative Analysis}
To qualitatively assess the logical validity of our method, we define one axiom and one premise:

\begin{itemize}
    \item \textbf{Axiom}: A correct verification process should not classify incorrect data as correct.
    \item \textbf{Premise}: Human inspectors make no mistakes during verification.
\end{itemize}

Figure~\ref{fig:cc_scenarios} illustrates the four main scenarios that arise when regenerating charts from captions:

\begin{enumerate}
    \item \textbf{Scenario A.} If the caption is incorrect, it produces faulty code leading to a mismatched image, which is identified and removed.
    \item \textbf{Scenario B.} If the caption lacks sufficient detail, an oversimplified chart is generated and subsequently filtered out.
    \item \textbf{Scenario C.} Even if the caption is accurate, errors in code generation or execution can result in a failed reconstruction, leading the sample to be excluded.
    \item \textbf{Scenario D.} Only when the caption is both accurate and informative, and the chart regenerates without errors, does the sample pass verification.
\end{enumerate}

This process ensures that only captions containing both correct and adequately detailed information are retained.

In summary, our process is designed to guarantee both the correctness and depth of information in \ChartCap, while substantially boosting the efficiency of large-scale verification. By combining logical verification with cycle consistency-based human verification process, we enable efficient quality control of \ChartCap and mitigate the burden of manual inspection.

\section{Validation of VCS with Human Evaluation}
\label{appendix:vcs_validation}
To validate the effectiveness of the Visual Consistency Score (VCS), we performed head‑to‑head human evaluations. For every comparison between two baselines, we randomly sampled 100 chart–caption pairs from the three test sets--\ChartCap, VisText, and Chart-to-Text. Following the protocol in Appendix~\ref{appendix:human_eval}, human annotators compared caption pairs and selected the better one with respect to informativeness, accuracy, and fewer hallucinations. We then computed the agreement rate as the proportion of comparisons in which the caption preferred by human annotators also received a higher metric score.

As shown in Table~\ref{fig:vcs_agreement}, VCS achieved the highest agreement rates across all three criteria, followed by OCRScore. These results indicate that both metrics reliably capture key aspects of caption quality as perceived by humans, validating their effectiveness as automatic metrics.

{\renewcommand{\arraystretch}{1.0}
\begin{table}[t!]
\centering
\begin{adjustbox}{width=0.47\textwidth}
\begin{tabular}{lccc}
\toprule
\textbf{Metric} & \textbf{Informativeness} & \textbf{Accuracy} & \textbf{Fewer hallucinations} \\
\midrule
SacreBLEU              & 60.50 & 59.50 & 59.83 \\
ROUGE             & 55.34 & 55.67 & 57.00 \\
METEOR            & 70.34 & 68.34 & 69.34 \\
BERTScore         & 68.67 & 67.00 & 68.34 \\
\midrule
VCS (so400m)      & \textbf{79.33} & \textbf{77.00} & \textbf{77.33} \\
OCRScore          & \underline{76.00} & \underline{75.00} & \underline{74.00} \\
\bottomrule
\end{tabular}
\end{adjustbox}
\caption{Agreement between automated metrics and human judgments (\%). Higher is better.}
\label{fig:vcs_agreement}
\end{table}}

\section{LLM Fidelity in Caption-to-Code Translation}
We observe that caption distortions during the first phase of VCS evaluation--LLM caption-to-code translation--are rare in practice. Interestingly, the distortion rate increases slightly when a caption is less informative. To investigate this, we analyzed 100 caption-code pairs each from Phi3.5-Vision-4B$_\text{\ChartCap}$ (with the highest VCS) and Phi3.5-Vision-4B$_\text{Original}$ (with the lowest VCS). We examined (1) whether the elements defined in the caption were correctly preserved, and (2) whether any content not described in the caption appeared in the code.

As a result, Phi3.5-Vision-4B$_\text{\ChartCap}$ achieved a caption-to-code accuracy of 99\%, with the remaining 1\% due to the omission of the axis title. For Phi3.5-Vision-4B$_\text{Original}$, the accuracy was 96\%, and in the remaining 4\% of cases, the LLM hallucinated placeholder or arbitrary data values to fill in the missing details of oversimplified captions. These results show that (1) translation errors are infrequent, and (2) lower information density in captions tends to increase the likelihood of LLM's caption-to-code distortions, ultimately resulting in lower VCS.

\section{Sensitivity of VCS to Structural Errors}
Although SigLIP's attention mechanism on charts is not fully interpretable, we find that the model is reasonably sensitive to structural elements. We analyze 100 captions collected from baseline models and found three major error types: (1) misidentification of maxima/minima (20\%), (2) axis hallucinations (13\%), and (3) omission of data series (7\%). After manually correcting these errors, VCS increased by 1.3\%, 6.1\%, and 4.7\%, respectively, indicating that VCS is capable of detecting such structural issues.

\section{Additional Baselines}
\label{appendix:baselines}
We additionally fine‑tuned Qwen2.5‑VL‑7B on \ChartCap and evaluated it on the VisText and Chart‑to‑Text benchmarks. We also evaluated Phi3.5-Vision-4B$_{\text{ChartSumm}}$ on the same benchmarks.

As shown in Table~\ref{tab:add_baseline}, Qwen2.5‑VL‑7B$_\text{\ChartCap}$ consistently outperforms its base model, whereas Phi3.5-Vision-4B$_\text{ChartSumm}$ underperforms Phi3.5-Vision-4B$_\text{\ChartCap}$ and even degrades performance relative to its own base model.

\begin{figure}
    \centering
    \includegraphics[width=\columnwidth]{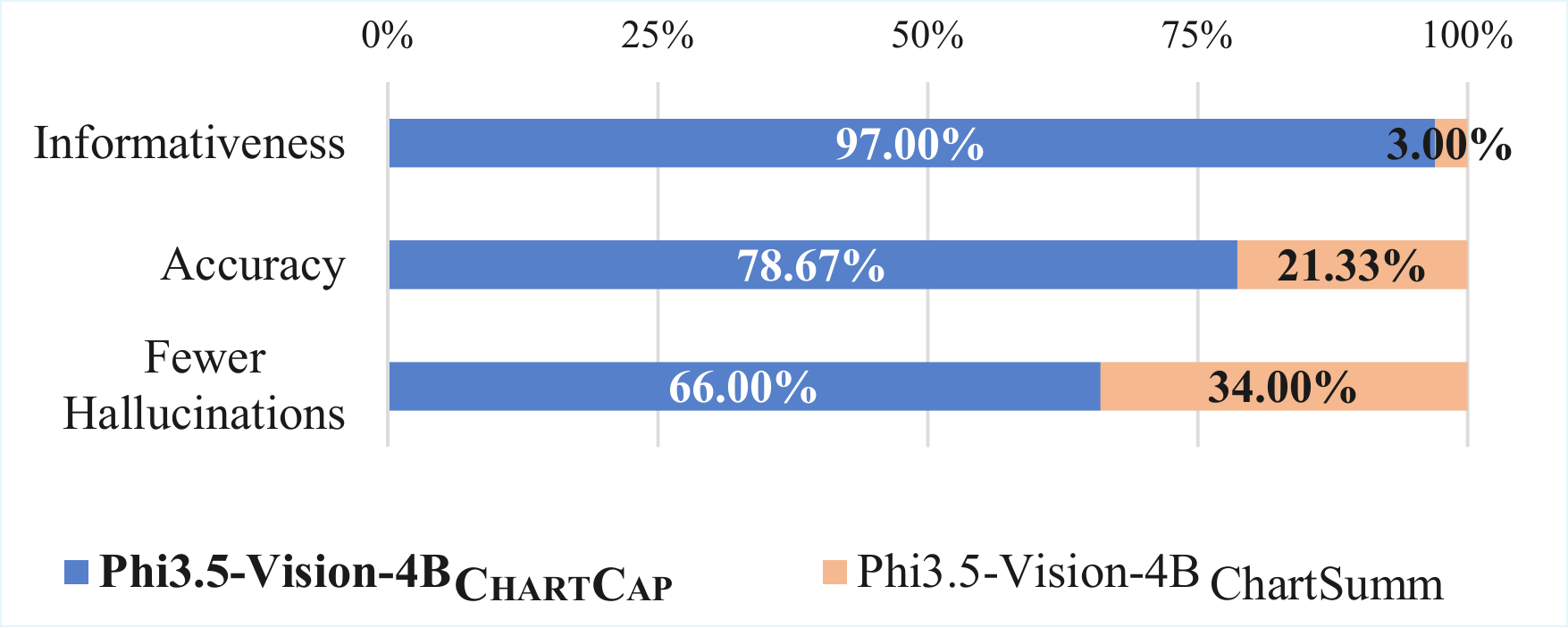}
    \caption{Human evaluation results comparing Phi3.5-Vision-4B$_{\text{\ChartCap}}$ and Phi3.5-Vision-4B$_{\text{ChartSumm}}$ on the VisText test set.}
    \label{fig:humaneval_chartsumm}
\end{figure}

We conduct additional human evaluation under the protocol in Appendix~\ref{appendix:human_eval} directly comparing Phi3.5-Vision-4B$_\text{\ChartCap}$ and Phi3.5-Vision-4B$_\text{ChartSumm}$ on VisText test set. As shown in Figure~\ref{fig:humaneval_chartsumm}, Phi3.5-Vision-4B$_{\text{ChartSumm}}$ received fewer preferences than Phi3.5-Vision-4B$_{\text{\ChartCap}}$ across all three evaluation aspects.
In summary, both automatic and human evaluations indicate that (1) \ChartCap consistently improves the captioning performance of state-of-the-art models, and (2) \ChartCap is a more effective training dataset than ChartSumm.

{\renewcommand{\arraystretch}{0.8}
\begin{table}[t]
\scriptsize
\centering
\begin{tabular}{lcccc}
\toprule
\textbf{Model} & \multicolumn{2}{c}{\textbf{VisText}} & \multicolumn{2}{c}{\textbf{Chart-to-Text}} \\
\cmidrule(lr){2-3} \cmidrule(lr){4-5}
& VCS & OCRScore & VCS & OCRScore \\
\midrule
Qwen2.5-VL-7B & 0.9044 & 0.3197 & 0.7739 & 0.1622 \\
\rowcolor{lightgray}Qwen2.5-VL-7B$_{\text{\ChartCap}}$ & 0.9328 & 0.3436 & \textbf{0.8084} & \textbf{0.1817} \\
\midrule
Phi3.5-Vision-4B & 0.8814 & 0.3414 & 0.7490 & 0.1786 \\
\rowcolor{lightgray}Phi3.5-Vision-4B$_{\text{\ChartCap}}$ & \textbf{0.9382} & \textbf{0.3826} & 0.8075 & 0.1789 \\
\rowcolor{lightgray}Phi3.5-Vision-4B$_{\text{ChartSumm}}$ & 0.8677 & 0.1414 & 0.7281 & 0.0789 \\
\bottomrule
\end{tabular}
\caption{Results of VCS and OCRScore on VisText and Chart-to-Text.VCS is computed using SigLIP2-So400M-512.}
\label{tab:add_baseline}
\end{table}
}

\section{Training Hyperparameters}
\label{appendix:hyperparameters}
Training was conducted using 6 RTX A6000 GPUs. The total training time was approximately 60 hours for InternVL2.5-8B$_{\text{\ChartCap}}$, 30 hours for Phi3.5-Vision-4B$_{\text{\ChartCap}}$, 12 hours for Phi3.5-Vision-4B$_{\text{Original}}$, 2 hours for Phi3.5-Vision-4B$_{\text{ChartSumm}}$, and 50 hours for Qwen2.5-VL-7B$_{\text{\ChartCap}}$. The training hyperparameters used for these models are summarized in Table~\ref{tab:train_settings}.

{\renewcommand{\arraystretch}{1.0}
\begin{table*}[t]
\centering
\begin{adjustbox}{max width=\textwidth}
\begin{tabular}{lccccc}
\toprule
\textbf{Hyperparameter} & \textbf{InternVL2.5-8B$_{\text{\ChartCap}}$} & \textbf{Phi3.5-vision-4B$_{\text{\ChartCap}}$} & \textbf{Phi3.5-vision-4B$_{\text{Original}}$} & \textbf{Phi3.5-vision-4B$_{\text{ChartSumm}}$} & \textbf{Qwen2.5-VL-7B$_{\text{\ChartCap}}$}\\
\midrule
Epochs        & 2 & 1 & 1 & 1 & 2     \\
Batch size    & 12 & 192 & 192 & 192 & 12    \\
Learning rate & 2e-5 & 2e-5 & 2e-5 & 2e-5 & 2e-5   \\
Optimizer     & AdamW & AdamW & AdamW & AdamW & AdamW \\
Warmup ratio  & 0.05    & 0.05  & 0.1 & 0.05 & 0.1 \\
Scheduler     & cosine & constant & cosine & constant & cosine \\
LoRA rank     & 32 & 32 & 32 & 32 &32      \\
LoRA alpha    & 64 & 32 & 32 & 32 &64      \\
LoRA dropout    & 0.0 & 0.05 & 0.05 & 0.05 & 0.05      \\
\bottomrule
\end{tabular}
\end{adjustbox}
\caption{Training hyperparameters for InternVL2.5-8B, Phi3.5-vision-4B, and Qwen2.5-VL-7B.}
\label{tab:train_settings}
\end{table*}}

\section{Details of Human Evaluation}
\label{appendix:human_eval}
\textbf{Sampling and Setup.} 
For each comparison, we randomly sampled 100 chart–caption pairs from two competing baselines. Crowd workers were shown the two pairs in a random left‑right order and asked to choose the better pair under three criteria:
\begin{enumerate}
    \item \textbf{Informativeness} – Does the caption adequately describe the chart’s structure and key insights (highlighted in green and blue in Fig~\ref{fig:chartcap_example})?  
    \item \textbf{Accuracy} – How faithfully does the caption reflect the chart’s structure and key insights?
    \item \textbf{Fewer Hallucinations} – Does the caption avoid information that cannot be inferred from the chart (highlighted in red in Fig~\ref{fig:chartcap_example})?  
\end{enumerate}
For the dataset‑level study in \S~\ref{human_eval:dataset} (\ChartCap vs ChartSumm), we added a fourth question--overall preference--to capture holistic quality while accounting for chart complexity.

\begin{figure*}[t]
\begin{center}
\includegraphics[width=0.67\textwidth]{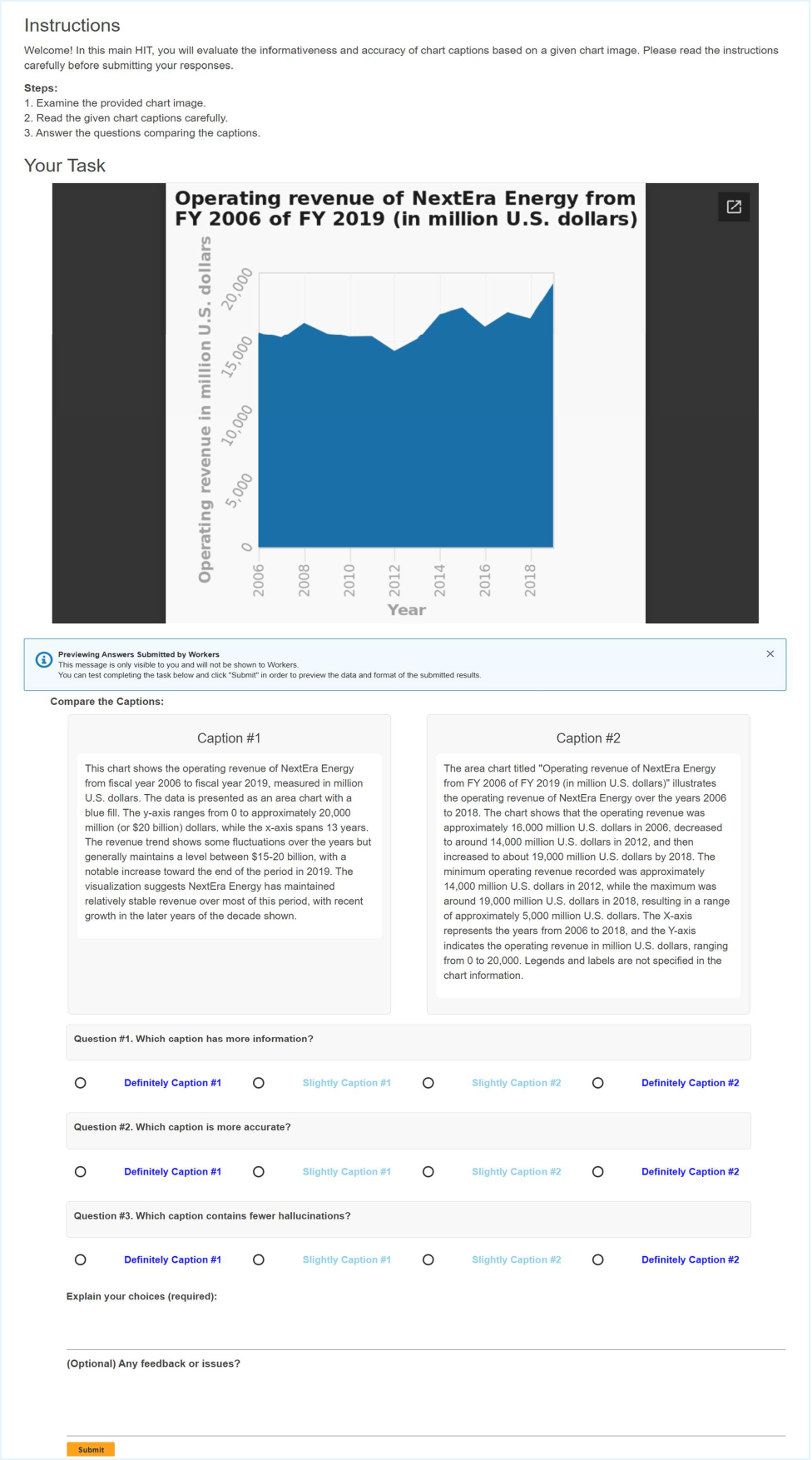}  
\end{center}
\caption{User interface for human evaluation comparing captions from different models on informativeness, accuracy, and fewer hallucinations.}
\label{fig:UI}
\end{figure*}

\begin{figure*}[t]
\begin{center}
\includegraphics[width=0.71\textwidth]{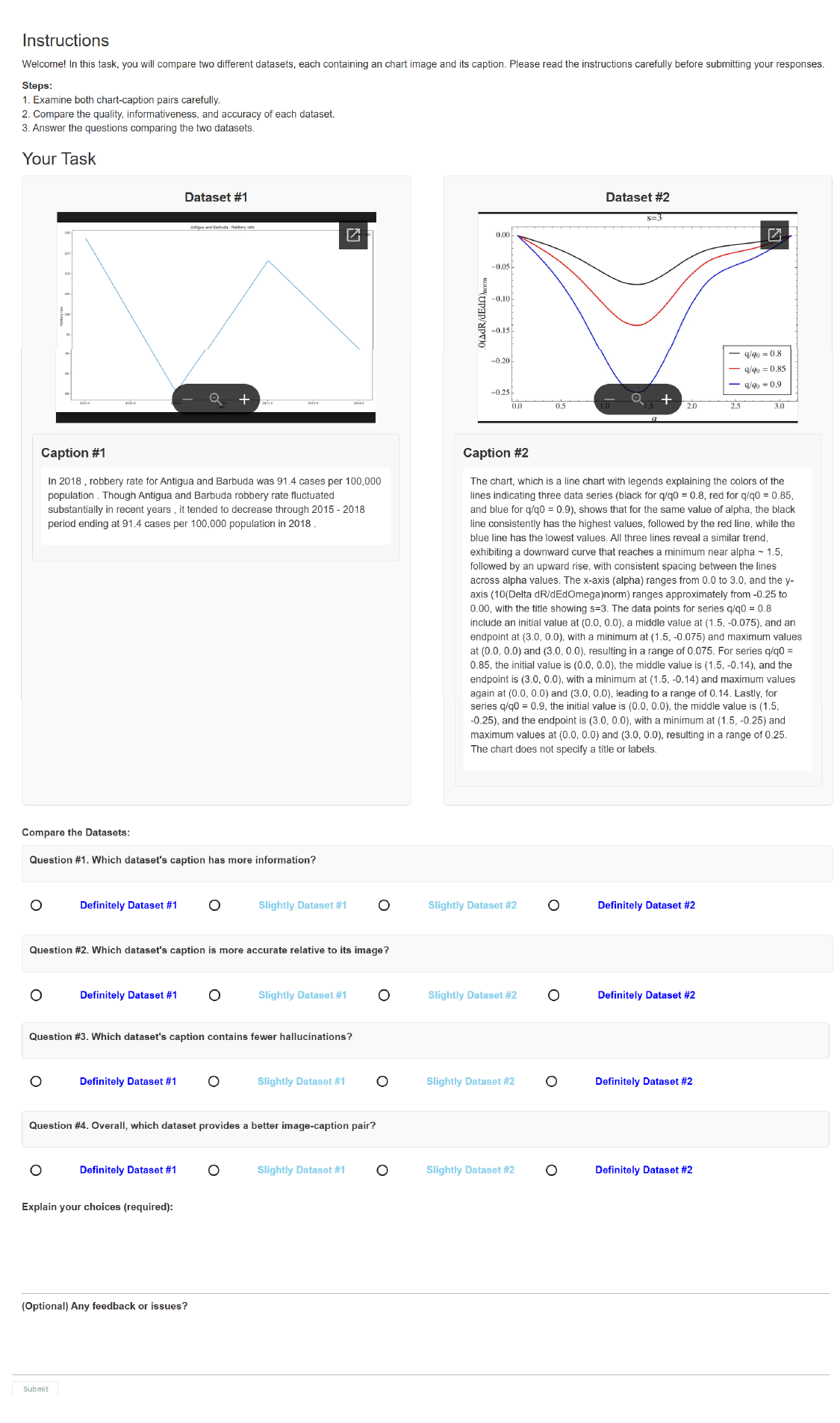}  
\end{center}
\caption{User interface for human evaluation comparing datasets (\ChartCap vs. ChartSumm) on informativeness, accuracy, fewer hallucinations, and overall preference.}
\label{fig:UI_dataset_comparison}
\end{figure*}

\textbf{User Interfaces.}
Figure~\ref{fig:UI} illustrates the user interface used for model comparisons, while Fig~\ref{fig:UI_dataset_comparison} shows the interface used for dataset comparisons.

\textbf{Worker Qualification and Quality Control.} 
To ensure reliable judgments, we administered a qualification test to assess workers’ understanding of the task. Only those who passed were allowed to participate in the main evaluation. During the evaluation, workers were also required to provide brief justification for their choices, discouraging random or inattentive responses.

\textbf{Platform and Demographics.}
Evaluations were conducted on Amazon Mechanical Turk, with participation restricted to English‑speaking countries (Australia, Canada, New Zealand, the United States, and the United Kingdom).

\textbf{Inter‑annotator agreement.} 
We measured the inter‑annotator agreement using Gwet’s AC1.  
\begin{itemize}
  \item \S~\ref{human_eval:dataset}: 0.84 (informativeness), 0.80 (accuracy), 0.23 (fewer hallucinations), 0.80 (overall preference).
  \item \S~\ref{human_evaluation:chartcap}: 0.47, 0.27, 0.27.
  \item \S~\ref{human_evaluation:vistext}: 0.70, 0.52, 0.22.
  \item \S~\ref{appendix:vcs_validation}: 0.84, 0.71, 0.71.
  \item \S~\ref{appendix:baselines}: 0.91, 0.37, 0.24
\end{itemize}

\textbf{Compensation.}
Workers were compensated \$0.50 per HIT, corresponding to approximately \$15 per hour, which exceeds the U.S. federal minimum hourly wage.

\end{document}